\pdfoutput=1

\documentclass[11pt]{article}

\usepackage[]{acl}

\usepackage{times}
\usepackage{latexsym}

\usepackage[T1]{fontenc}

\usepackage[utf8]{inputenc}

\usepackage{microtype}
\usepackage{hyperref}
\usepackage{url}
\usepackage{newtxmath} 
\usepackage{amsmath}
\usepackage{mathtools}
\usepackage{tabularx}
\usepackage{todonotes}
\usepackage{multirow}
\usepackage{enumitem}
\usepackage{hhline}
\usepackage{booktabs}
\usepackage{subcaption}
\usepackage{stfloats}
\usepackage{colortbl} 
\usepackage{soul} 
\usepackage{ulem}
\usepackage{cancel}

\definecolor{bluegrey}{RGB}{230, 240, 255}
\definecolor{lightgrey}{HTML}{dcdbdb}
\sethlcolor{bluegrey} 
\newcommand{\cc}[0]{\cellcolor{bluegrey}}
\newcommand{\dd}[0]{\cellcolor{lightgrey}}
\definecolor{green}{RGB}{0,128,0}
\definecolor{purple}{RGB}{75,0,130}
\definecolor{add}{RGB}{248, 206, 204}
\definecolor{forestgreen}{RGB}{34,139,34}
\definecolor{grey}{RGB}{238, 238, 238}
\definecolor{textorange}{RGB}{255,140,0}
\definecolor{textblue}{RGB}{0,128,255}
\definecolor{textgrey}{RGB}{128,128,128}
\definecolor{darkred}{RGB}{196, 30, 58}
\definecolor{bluegray}{rgb}{0.4, 0.6, 0.8}
\definecolor{bluencs}{rgb}{0.0, 0.53, 0.74}
\definecolor{newblue}{RGB}{0,102,204}

\newcommand{\addinfo}[0]{\textcolor{textorange}}
\newcommand{\keepinfo}[0]{\textcolor{newblue}}
\newcommand{\rv}[1]{\mathsf{#1}}

\DeclarePairedDelimiterX{\infdivx}[2]{(}{)}{%
  #1\;\delimsize\|\;#2%
}
\newcommand{\infdiv}{\textit{KL}\infdivx}

\newcommand{\defeq}{\mathrel{\stackrel{\textnormal{\tiny def}}{=}}}

\DeclarePairedDelimiter\abs{\lvert}{\rvert}

\usepackage{lipsum}
\makeatletter
\newcommand{\balancecolsandclearpage}{
  \close@column@grid
  \cleardoublepage
  \twocolumngrid
}
\makeatother

\newcommand{\qt}[1]{\textcolor{orange}{\bf \small [#1 --qt]}}

\title{Explanation Regeneration via Information Bottleneck}

\author{
Qintong Li$^{\spadesuit}$\thanks{\, Work done while interning at Tencent AI Lab.}$\;\,$
\textbf{Zhiyong Wu}$^{\diamondsuit}$
\textbf{Lingpeng Kong}$^{\spadesuit}$
\textbf{Wei Bi}$^{\heartsuit}$
\\
$^\spadesuit$The University of Hong Kong \\
$^\diamondsuit$Shanghai AI Laboratory \quad
$^\heartsuit$Tencent AI Lab \\
{\small
{ \fontfamily{txtt}\selectfont qtli@connect.hku.hk, wuzhiyong@pjlab.org.cn, lpk@cs.hku.hk, victoriabi@tencent.com}
}
}

\begin{document}
\maketitle
\begin{abstract}
Explaining the black-box predictions of NLP models naturally and accurately is an important open problem in natural language generation.
These free-text explanations are expected to contain sufficient and carefully-selected evidence to form supportive arguments for predictions. 
Thanks to the superior generative capacity of large pretrained language models (PLM), recent work built on prompt engineering enables explanations generated without specific training. However, explanations generated through single-pass prompting often lack sufficiency and conciseness, due to the prompt complexity and hallucination issues.
To discard the dross and take the essence of current PLM's results, we propose to produce sufficient and concise \underline{\textbf{e}}xplanations via the \underline{\textbf{i}}nformation \underline{\textbf{b}}ottleneck (\texttt{EIB}) theory. \texttt{EIB} regenerates explanations by polishing the single-pass output of PLM but retaining the information that supports the contents being explained by balancing two information bottleneck objectives. 
Experiments on two different tasks verify the effectiveness of \texttt{EIB} through automatic evaluation and thoroughly-conducted human evaluation. 
\end{abstract} 

\section{Introduction}

Natural language explanations have attracted a lot of attention as a way to uncover the rationales behind black-box predictions.
Thanks to the power of large pretrained language models (PLM)~\citep{gpt-3,opt}, prompting methods proposed in recent studies achieve impressive results in generating free-text explanations~\citep{WeiWSBCLZ22,LampinenRDCTMYS22}. 
A clear advantage of such methods is that they involve no additional training from task-specific datasets.

\begin{figure}[!t]
    \centering
\includegraphics[width=0.47\textwidth]{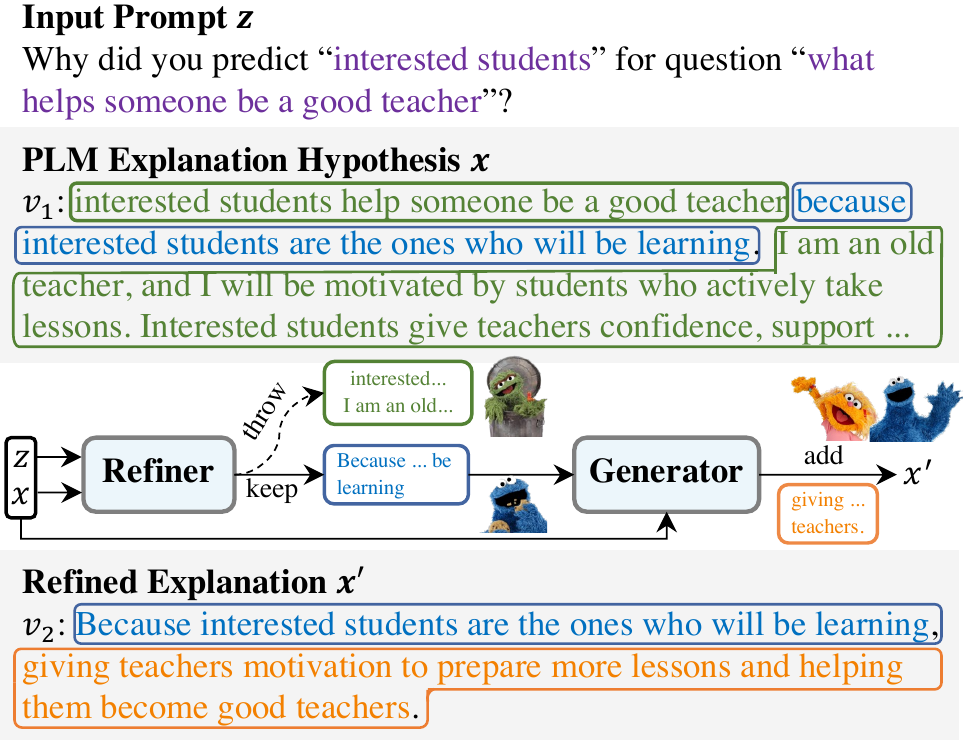}
    \caption{Although PLM generates an informative explanation hypothesis ($v_1$), this explanation contains \textcolor{green}{redundant} or \textcolor{green}{inessential information} which may interfere with the holistic understanding of the relationship between \textcolor{purple}{question} and \textcolor{purple}{answer}. In comparison, the polished explanation ($v_2$), improved upon the initial hypothesis, is more concise and reasonable.}

\label{fig:intro_case}
\end{figure}

In this paper, 
we regard a free-text explanation as a description of the relationship between an input context and a hypothesis, e.g., a question and an answer.
Although it is difficult to state that one explanation is superior to all others due to the different desiderata of the tasks to be explained, this does not prevent us from answering the question ``\textit{what makes a good explanation}'' from a practical view. 
Previous research~\citep{YuCZJ19,Miller19} points out several semantic constraints should be satisfied in constructed explanations: (i) avoid undesirable content, like repeating context's statement, (ii) ensure adequate background supports, and (iii) emphasize selective evidence. 
Current machine-generated explanations still exhibit defects on these constraints \citep{kassner-schutze-2020-negated,welleck2022generating}. 
For single-pass prompting methods, they cast the burden of ensuring explanation constraints all on a PLM which ``starts from scratch''.
This inspires us to investigate how to discard the dross and take the essence of current PLM's results.

We propose our \underline{\textbf{e}}xplanation generation approach via the \underline{\textbf{i}}nformation \underline{\textbf{b}}ottleneck theory~\citep{Tishby2000} (\texttt{EIB}), which can refine explanations prompted from PLM into more \textit{meaningful}, \textit{sufficient}, and \textit{concise} ones.
It works in two phases, as illustrated in Figure~\ref{fig:intro_case}.
First, given an NLP task sample (e.g., a QA pair), \texttt{EIB} uses a large PLM to produce an initial explanation hypothesis ($v_1$) by framing the task sample into a prompt input. 
Second, a \textit{refiner} improves the quality of an explanation hypothesis along the axis of the aforementioned characteristics (i.e., meaningful, sufficient, and concise).
The \textit{refiner} is trained following the information bottleneck principle. 
Concretely, it learns a minimal sufficient bottleneck representation of the explanation $v_1$, 
while being maximally explainable about the sample (i.e., the QA pair) by introducing an information loss~\citep{EthayarajhCS22}. With the learned bottleneck representation on hand, a generator learns to produce a new explanation.
We propose a simple and general procedure for training the refiner by pairing synthetic explanation hypotheses with gold references from existing datasets. 
\texttt{EIB} is a general explanation generation framework and can be applied to different NLP tasks with no specific task supervision.

We demonstrate the effectiveness of \texttt{EIB} in generating explanations on two popular NLP tasks: commonsense question answering and natural language inference. 
Experiments show that \texttt{EIB} significantly improves the explanation candidates prompted from PLM, by making them more concise while retaining useful information for explaining task samples. 
Automatic evaluation and carefully designed human evaluation demonstrate the performance of \texttt{EIB}. 
Furthermore, an analysis of evaluations shows an imperious demand for better metrics to judge explanations more credibly.
We publicly release our code and data\footnote{\url{https://github.com/qtli/EIB}}.

\section{Method}
\label{sec:method}

\paragraph{Prompting} 
Recently, writing explanations through prompting large PLMs has become a competitive approach. Given an NLP task sample $\boldsymbol{z}$ including input $\boldsymbol{z}_c$ and output $\boldsymbol{z}_o$, we could infer its explanation $\boldsymbol{x}$ via prompting a PLM:  $\boldsymbol{x}=\textit{PLM}(S(\boldsymbol{z}_c, \boldsymbol{z}_o))$, where function $S(\cdot,\cdot)$ transforms $\boldsymbol{z}$ to prompt formats through predefined templates. For example, if we have a QA sample, question $\boldsymbol{z}_c$: \textit{Can elephants be put in the fridge?} and answer $\boldsymbol{z}_o$: \textit{no}, the prompt will be ``\textit{The question is can elephants be put in the fridge? The answer is no \texttt{because}.}''.

Although prompting has achieved remarkable success, machine-generated explanations still have room for improvement as discussed in the introduction.
Therefore, we seek to step further under the current achievement, exploring an effective way to improve explanation quality in terms of meaningfulness, sufficiency, and conciseness.

\paragraph{Formulation}
Suppose we have a sample $\boldsymbol{z}\in \rv{z}$ and its explanation hypothesis $\boldsymbol{x}\in \rv{x}$.\footnote{$\boldsymbol{x}$,$\boldsymbol{t}$,$\boldsymbol{z}$ and $\mathbf{X}$,$\mathbf{T}$,$\mathbf{Z}$ are instances of random variables $\rv{x}$,$\rv{t}$,$\rv{z}$.}  
We aim to refine $\boldsymbol{x}$ into a better $\boldsymbol{x}{'}$ which can: (1) reduce irrelevant information in $\boldsymbol{x}$ (conciseness), (2) preserve and supplement useful information to infer $\boldsymbol{z}$ (meaningfulness, sufficiency).
We divide the explanation regeneration task into two problems: \textit{refinement} and \textit{generation}.

First, we model the refinement problem from an information-theoretic view, i.e., learn the internal representation $\rv{t}$ of the initial explanation $\rv{x}$, defined as $p_\theta(\boldsymbol{t}\mid\boldsymbol{x})$, such that $\rv{t}$ is maximally compressive about the (noisy) $\rv{x}$ while being maximally expressive about $\rv{z}$:
\begin{align}
\min_\theta \mathrm{I}(\rv{x},\rv{t}) \textsf{ s.t. } \mathrm{I}(\rv{t},\rv{z}) \geq \mathrm{I}_c \, ,
\end{align}
The above process can be formulated as the \textbf{information bottleneck principle} (IB)~\citep{TishbyZ15,AlemiFD017}. IB defines the characteristics of an optimal representation, in terms of the fundamental tradeoff between having a concise representation and one with good predictive power, which is equivalent to minimizing the following objective function:

\begin{align}
\mathcal{L}_{\textit{IB}} = \beta \cdot\underbrace{\mathrm{I}(\rv{x},\rv{t})}_{\textit{compression}}-\underbrace{\mathrm{I}(\rv{t},\rv{z})}_{\textit{preservation}} \, .
\label{eq:IB}
\end{align} 
where $\beta$ is a Lagrange multiplier. A large $\beta$ corresponds to high compression, and hence low mutual information between $\rv{t}$ and $\rv{z}$.

\begin{figure*}[!t]
    \centering
    \includegraphics[width=\textwidth]{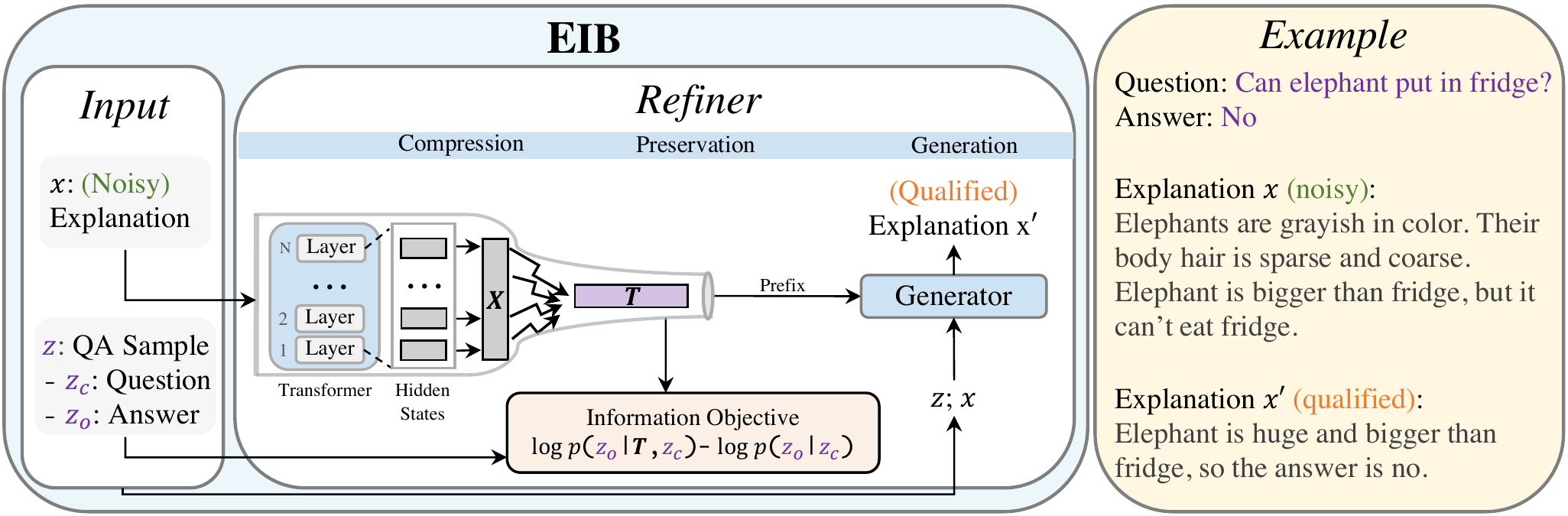}
    \caption{Illustration of our method. Given a task sample $\mathrm{z}$ and an explanation candidate $\boldsymbol{x}$ which may be noisy, (i) a refiner first compresses $\boldsymbol{x}$ into bottleneck vectors $\mathbf{T}$ via a tunable stochastic mapping. (ii) An information objective optimizes compression direction ensuring $\mathbf{T}$ to be predictive of $\boldsymbol{z}$. (iii) A generator generates a sufficient and concise explanation based on the bottleneck representation $\mathbf{T}$, $\boldsymbol{z}$, and $\boldsymbol{x}$. 
    The right side shows an example of \texttt{EIB}.} 
    \label{fig:overall_framework} 
\end{figure*}

Given a bottleneck representation $\boldsymbol{t}$, our second goal is to generate a free-text explanation $\boldsymbol{x}{'}$ based on $\boldsymbol{x}$.
Therefore, we pack a log-likelihood objective for language modeling with $\mathcal{L}_{\textit{IB}}$ as the objective function of the whole model, and train it on an automatically constructed synthetic dataset:
\begin{align}
 \mathcal{L}_{\textit{EIB}} = \underbrace{\mathcal{L}_{\textit{IB}}}_{\textit{refinement}}- \,\underbrace{\log p(\boldsymbol{x}{'}\mid\boldsymbol{t},\boldsymbol{x},\boldsymbol{z})}_{\textit{generation}} \, . 
\label{eq:EIB}
\end{align} 
The overall proposed \texttt{EIB} is illustrated in Figure~\ref{fig:overall_framework}.

In the following, we will present the optimization and training with respect to (i) explanation compression for distilling a bottleneck representation from the initial explanation, (ii) information preservation for ensuring the distilled bottleneck representation expressive about the explained sample, and (iii) explanation regeneration from the distilled bottleneck representation for producing a better explanation than the initial input one.

\subsection{Explanation Compression}
\label{sec:compression}

\paragraph{Vectorization}
Suppose we have an explanation candidate $\boldsymbol{x}$ that needs to be improved. We first use a parameter-fixed $n$-layer PLM to encode $\boldsymbol{x}$ and aggregate the hidden states of $n$ layers into a sequence of vectors $\mathbf{X}\in\mathbb{R}^{n\times d}$, where each $d$-dimensional vector $\mathbf{x}_i$ is the weighted sum of hidden representations of the corresponding layer by attention weights.
We utilize representations of all layers instead of the last layer only in order to combine more information.

\paragraph{Compression}
Our first goal is to denoise irrelevant information in $\mathbf{X}$ and obtain a highly compact representation $\mathbf{T}$. The compression loss part in $\mathcal{L}_{\textit{IB}}$ can be rewritten as:
\begin{align}
    \mathrm{I}(\rv{x};\rv{t})\defeq \sum_i^n \mathbb{E}_{\mathbf{x}_i}[\mathbb{E}_{\mathbf{t}_i\sim p_{\theta}}[\log(\frac{p_{\theta}(\mathbf{t}_i\mid\mathbf{x}_i)}{p_{\theta}(\mathbf{t}_i)})]] \, ,
\end{align}
where $p_\theta(\mathbf{t}_i)$ is the prior distribution of the bottleneck vector $\mathbf{t}_i$, $p_{\theta}(\mathbf{t}_i\mid\mathbf{x}_i)$ is the stochastic mapping from the distribution of initial explanation hypothesis to its intermediate compressed representation, and $\theta$ indicates learnable parameters. 

\paragraph{Optimization} 
Specifically, we perform a linear transformation on each vector $\mathbf{x}_i$ of $\mathbf{X}$, to produce a polished representation $\mathbf{T}=\textit{MLP}(\mathbf{X})\in\mathbb{R}^{n\times k}$. 
We assume each vector $\mathbf{t}_i$ of $\mathbf{T}$ follows an isotropic Gaussian distribution, where the mean and standard deviation are learnable parameters with the use of the reparameterization trick.
However, for $p_{\theta}(\mathbf{t}_i)=\mathbb{E}_{\hat{\mathbf{x}_i}}[p_{\theta}(\mathbf{t}_i\mid\hat{\mathbf{x}}_i)]$,  it is difficult to loop over all candidates $\hat{\mathbf{x}_i}$. We practically use a standard Gaussian distribution $p_{\mathcal{N}}(\mathbf{t}_i) \sim \mathcal{N}(\boldsymbol{0}, \boldsymbol{1})$ to simulate $p_{\theta}(\mathbf{t}_i)$ for simplicity.
Using the fact $\mathbb{E}[\infdiv{p_\theta(\mathbf{t}_i)}{p_{\mathcal{N}}(\mathbf{t}_i)}] \geq 0$, we can minimize the upper bound of $\mathrm{I}(\rv{x};\rv{t})$:
\begin{align}
   \mathrm{I}(\rv{x},\rv{t}) \leq \sum_i^n\mathbb{E}_{\mathbf{x}_i}[\mathbb{E}_{\mathbf{t}_i\sim p_{\theta}}[\log(\frac{p_{\theta}(\mathbf{t}_i|\mathbf{x}_i)}{p_{\mathcal{N}}(\mathbf{t}_i)})]]\, .
\end{align}
Making the bound as tight as possible given $\theta$ allows yielding a compressed representation $\mathbf{T}$ distilled from the initial $\mathbf{X}$.

\subsection{Information Preservation}
\label{sec:t-info}

The second goal of IB in Eq.~\ref{eq:IB} is to maximize $\mathrm{I}(\rv{t},\rv{z})$,
which can lead to a high log-likelihood $p_\theta(\mathbf{Z}\mid\mathbf{T})$ for ensuring $\mathbf{T}$ not losing predictive features of $\mathbf{X}$ to explain $\mathbf{Z}$:
\begin{align}
\small
    \mathrm{I}(\rv{t},\rv{z})&\defeq\sum_i^n\mathbb{E}_{\mathbf{z}_i,\mathbf{t}_i\sim p_{\theta}}[\log(\frac{p_{\theta}(\mathbf{z}_i\mid\mathbf{t}_i)}{p(\mathbf{z}_i)})]\, ,\\
    p_{\theta}(\mathbf{z}_i\mid\boldsymbol{t}_i)&\defeq\sum_i^n\mathbb{E}_{\mathbf{x}_i}[\frac{p(\mathbf{z}_i\mid\mathbf{x}_i)p_{\theta}(\mathbf{t}_i\mid\mathbf{x}_i)p(\mathbf{x}_i)}{p_{\theta}(\mathbf{t}_i)}]\, .
\end{align}
However, $p_{\theta}(\mathbf{z}_i\mid\mathbf{t}_i)$ is hard to estimate because we have to iterate on all possible $\boldsymbol{x}_i$. Furthermore, the length of $\boldsymbol{z}$ is not fixed and cannot be precisely aligned to the number of bottleneck vectors $\mathbf{T}$.

\paragraph{Optimization}
We extend recent work in information theory~\citep{xuinfo20,EthayarajhCS22}, {which generalizes Shannon's information theory to quantify the predictive $\mathcal{V}$-information between two random variables, subject to computational constraints $\mathcal{V}$.  $\mathcal{V}$-information reflects the ease with which $\mathcal{V}$ can predict $\rv{z}$ given $\rv{t}$.

In this paper, we use $p_\phi$ to denote the computational constraints, i.e., an autoregressive model GPT-2~\citep{radford2019language}. Measuring $\mathrm{I}(\rv{t},\rv{z})$ becomes quantifying usable information under $p_\phi$.
Then $\mathrm{I}(\rv{t},\rv{z})$ can be approximated by the information difference of an unconditional entropy $H_{p_\phi}(\rv{z})$ and conditional entropy $H_{p_\phi}(\rv{z}\mid\rv{t})$ \textit{w.r.t} computation-bounded parameters $\phi$:
\begin{align}     
\mathrm{I}(\rv{t},\rv{z}) & \geq H_{p_\phi}(\rv{z}) - H_{p_\phi}(\rv{z}\mid\rv{t})\, ,\\
H_{p_\phi}(\rv{z}) &=\mathbb{E}_{\boldsymbol{z}}[- \log p_\phi(\boldsymbol{z}) ]\, ,\\ 
H_{p_\phi}(\rv{z}\mid\rv{t}) &= \mathbb{E}_{\boldsymbol{z},\mathbf{T}\sim p_\theta(\mathbf{T}\mid\mathbf{X})}[- \log p_\phi(\boldsymbol{z}\mid\mathbf{T})]\, ,
\end{align}
where $\theta$ and $\phi$ are optimizable parameters, $\mathbf{t}$ acts as a learnable prefix~\citep{LiL20} to a GPT-2. 

Optimizing the lower bound of $\mathrm{I}(\rv{t},\rv{z})$
\begin{align*}
    \mathbb{E}_{\boldsymbol{x},\boldsymbol{z}}[\mathbb{E}_{\mathbf{T}\sim p_{\theta}(\mathbf{T}\mid\mathbf{X})}[\log p_{\phi}(\boldsymbol{z}\mid\mathbf{T})-\log p_\phi(\boldsymbol{z})]]
\end{align*}
requires $\mathbf{T}$ to have enough capacity to support $\boldsymbol{z}$ while being compact with the consideration of the minimization of $\mathrm{I}(\rv{x},\rv{t})$.

\subsection{Explanation Regeneration}
\label{sec:expl-gen}

With the distilled bottleneck representation $\mathbf{T}$ on hand, the remaining task is to translate the compact representation into a new explanation $\boldsymbol{x}{'}$ that may be different from the initial explanation $\boldsymbol{x}$ while achieving obvious quality improvements.

Translating the highly-dimensional matrix $\mathbf{T}$ into a discrete and readable explanation is not an easy task. 
To tackle this challenge, we use the explanation datasets from various NLP tasks and build a training corpus by pairing the human-written explanation with its synthetic imperfect version, which allows us to train \texttt{EIB} on the explanation regeneration task. 
Finally, for generating a new explanation autoregressively, a generator (GPT-2) is optimized by a language modeling loss: $\log p_{\delta}(\boldsymbol{x}{'}|\mathbf{t},\boldsymbol{x},\boldsymbol{z})$ where $\mathbf{t}$ serves as a learnable prefix input.

\subsection{Training Dataset Construction.}
\label{sec:data_construct}
Now we detail the automatic construction of the training dataset for optimizing \texttt{EIB}.
After analyzing the explanations generated by the state-of-art models~\citep{opt,gpt-3}, compared to humans, machines could be further improved in generating informative explanations with adequate rationales in fewer words, especially when prompts are long and complex.

\begin{table}[!t]
\centering
\small
\begin{tabular}{p{0.95\linewidth}}
\toprule
\textbf{Sample $\boldsymbol{z}$} \\
\quad \textbf{$\boldsymbol{z}_c$:} There are two statements and select which one is true. \textit{<s>} Sentence 1 is people get dry while taking a shower. Sentence 2 is people get wet while taking a shower.  \\
\quad \textbf{$\boldsymbol{z}_o$:} Sentence 2 is true.  \\
\textbf{Synthetic $\boldsymbol{x}$:} It is also said that the high level of chlorine in the water will make people wet while taking a shower or a bath.  (\textcolor{textgrey}{\textit{sentence-level replacement, span-level infilling}}) \\
\textbf{Target $\boldsymbol{x}{'}$:}  Water make people wet while taking a shower. \\
\midrule[0.03em]
\textit{Source}: Sen-Making~\citep{wang2019-make} \\
\bottomrule
\end{tabular}
\caption{An example of the constructed \textsc{MixExpl} dataset. Explanation hypothesis $\boldsymbol{x}$ is synthesized by two \textcolor{textgrey}{operations} based on the target explanation $\boldsymbol{x}$.} 
\label{tab:mixexpl_case}
\end{table}

We construct a synthetic training corpus \textsc{MixExpl} according to the generation characteristics of PLM. We choose six existing free-text explanation datasets across various NLP tasks: science QA~\citep{jansen2016s}, fact-checking~\citep{alhindi2018,kotonya2020explainable}, commonsense validation~\citep{wang2019-make}, and defeasible natural language inference~\citep{brahman2021learning}.

Specifically, for each gold explanation $\boldsymbol{x}{'}$ of six tasks, we randomly choose 2, 3, or 4 types from five operations on ground truth $\boldsymbol{x}{'}$ to get $\boldsymbol{x}$, which is guided by explanation properties expected to learn. 
For information, we have token- and sentence-level repetition.
For sufficiency, we do token- and sentence-level replacement, negation, and shuffle. 
For conciseness, we conduct span- and sentence-level infilling.

\begin{itemize}[wide=0\parindent,noitemsep,topsep=0em]

\item {Repetition}: Redundant texts need to be avoided in explanation texts. For a good explanation,  we either repeat an $N$-gram ($N$=1,2,3,4) in a random sentence or randomly select a sentence to repeat.

\item {Replacement}: Using irrelevant token spans or sentences will cause explanations wrongly describe the expected rationales. We replace random 15\% keywords in a random explanation sentence with their antonyms or randomly replace an explanation sentence with another one sampled from the rest of the gold explanations.

\item {Negation}: Negation words are crucial for accurately explaining without conflicting with the task sample in context. We perform negation alteration by adding or removing negation words for randomly-selected verbs of the explanations using rules defined in~\citep{GuanH20}.

\item {Shuffle}: Temporal causal relationship plays a crucial role in clearly and logically explaining. We randomly reorder the sentences of an explanation to create logical issues.

\item {Infilling}: The selection of crucial evidence relevant to the task at hand facilitates the generation of concise explanations. 
We augment the gold explanation with relevant but inessential contents by retrieving similar sentences from other explanations using Contriever~\citep{contriever} or expanding an explanation sentence with GLM~\citep{DuQLDQY022}.
\end{itemize}

Finally, we build a training corpus \textsc{MixExpl} of tuples (task sample, synthetic explanation, and gold explanation), and train \texttt{EIB} on \textsc{MixExpl}. Table~\ref{tab:mixexpl_case} displays an instance of \textsc{MixExpl} corpus.

During inference, given an NLP sample (it could be from any NLP task, even not belonging to $\mathcal{D}_{\abs{n}}$) and a prompt suffix like \texttt{because}, we first use PLM to generate an initial explanation hypothesis $\boldsymbol{x}$.
Then we use the trained \texttt{EIB} framework to produce a new explanation towards sufficiency and conciseness.
The prompting formats and examples are illustrated in Appendix~\ref{apx:prompt_format} table~\ref{tab:prompt_format}.

\section{Experiment}

\subsection{Experiment Setup}
Our experiments are organized into three sets:
We first evaluate the quality of explanations generated by \texttt{EIB} on different tasks and compare various baselines without explicit refinements towards sufficiency and conciseness~(\S\ref{sec:main_human}). 
We further analyze the performance improvement brought by the information bottleneck with training on synthetic dataset \textsc{MixExpl}~(\S\ref{sec:exp_ib}).
Lastly, we qualitatively assess the current development of explanation generation and the challenges for evaluation~(\S\ref{sec:qualitative_exp}).

\paragraph{Human Evaluation Metrics}
Human evaluation has very high priorities for open-ended text generations~\citep{ZhangKWWA20,Goyal22,LiLBRLK22}, and the explanation generation task is not exempt.
From the free-text language aspect, we  evaluate
(i) Grammaticality and 
(ii) Factuality. 
From the open-ended explanation aspect, we measure:
(iii) New Information, i.e., being informative and diverse instead of repeatedly copying the given context.
(iv) Sufficiency, i.e., answering ``\textit{why this [output] is assigned to this [input]}'' and stating the relationship between them.  
(v) Conciseness. i.e., being selective and comprehensive, not enumerating the complete set~\citep{YuCZJ19,wiegreffe2021teach}. 
Three crowd-sourced annotators are instructed to conduct comparisons for 200 samples of two NLP tasks. Average Krippendorff’s alpha is reported to indicate the inter-annotator agreement. 
More details of metrics and annotation pipelines are included in Appendix~\ref{apx:annotation}.

\begin{table}[!t]
\centering
\resizebox{\columnwidth}{!}{
\begin{tabular}{llccc}
    \toprule
    \textbf{Stage} & \textbf{Datasets} & \textbf{Training} & \textbf{Validation} & \textbf{Testing}  \\
    \midrule
   \dd{Training} & \dd\textsc{MixExpl}  & \dd6,848 &  \dd764 & \dd828 \\
   & - ScienceQA & ~~~665 & ~~82 & 101 \\
   & - Sen-Making & 1,329 & 174 & 177\\
   & - LIAR-PLUS  & 2,028 & 245 & 239\\
   & - PubHealth & 1,320 & 150 & 177\\
   & - E-$\delta$-NLI & 1,506 & 113 & 134\\
    \midrule
    \dd{Inference} & \dd{ECQA} & \dd- & \dd- & \dd2,194 \\
    & \dd{e-SNLI} & \dd- & \dd- & \dd9,184\\
    \bottomrule
\end{tabular}
}
\caption{Statistics of training and inference datasets.} 
\label{tab:stat_dataset}
\end{table}

\paragraph{Automatic Metrics}
We include reference-based metrics BLEU-$\mathit{n}$~\citep{PapineniRWZ02}, Rouge-$\mathit{n}$~\citep{lin2002manual} CIDEr~\citep{VedantamZP15} and  BERTScore~\citep{ZhangKWWA20} and diversity metric Distinct-$\mathit{n}$~\citep{LiGBGD16}. Besides, we measure the proportion of distinct tokens (Novelty) in explanation that do not occur in given task sample. We report the average length (AVGLEN) of explanations to provide hints on conciseness.

\begin{table*}[t!]
\centering
\resizebox{1.92\columnwidth}{!}{
\begin{tabular}{llccccccccc}
    \toprule
    \textbf{Datasets} & \textbf{Methods} & \textbf{Grammar} & \textbf{Factuality} & \textbf{New Information} & \textbf{Sufficiency} & \textbf{Conciseness} & $\alpha$ \\
    \midrule
    ECQA & Human  & 2.99 & 3.00 & 2.88 & 2.83 &  2.60 & 0.365 \\
    & \textsc{Supervised} &  2.94 & 2.86 & 2.52 & 2.40 & 1.84 & 0.439 \\
    \midrule[0.03em]
     & \textsc{BottleSum} &  1.95 & 2.67 & 2.26 & 1.57 & 1.75 & 0.411 \\ 
     & \textsc{Prompting} & 2.88 & 2.66 & 2.69 & 2.02 & 1.73 & 0.563 \\ 
     & \textsc{Prompting}-Filter & 2.90 & 2.81 & 2.64 & 2.30 & 1.77 & 0.668 \\
     & \textsc{Prompting}-\texttt{EIB} & ~~\cc2.97$\ddagger$ & ~~\cc2.79$\dagger$ & \cc\textbf{2.76} & ~~\cc2.17$\dagger$ & ~~\cc2.59$\ddagger$ & 0.393 \\ 
     & \textsc{Prompting}-Filter-\texttt{EIB} & \cc2.93 & \cc\textbf{2.82} &  ~~\cc2.74$\dagger$ &  ~~\cc\textbf{2.35}$\dagger$ & ~~\cc\textbf{2.56}$\ddagger$ & 0.449\\
     \midrule
    e-SNLI & Human & 2.96 & 2.93 & 2.97  & 2.79 &  2.88 & 0.363 \\
     & \textsc{Supervised} & 2.94  & 2.54 & 2.80 & 2.25 & 2.52 & 0.576 \\
    \midrule[0.03em]
     & \textsc{BottleSum} & 1.95  & 2.35 & 2.26  & 1.51 & 1.37 & 0.421 \\
     & \textsc{Prompting} & 2.97  & 2.21 & 2.72  & 1.85 & 1.23 & 0.615 \\
     & \textsc{Prompting}-Filter & 2.97 & 2.46 & 2.61 & 1.83 & 1.30 & 0.591 \\
    & \textsc{Prompting}-\texttt{EIB} & \cc2.98 & ~~\cc2.57$\ddagger$ & ~~\cc\textbf{2.84}$\dagger$ & ~~\cc\textbf{2.09}$\ddagger$ & ~~\cc\textbf{2.22}$\ddagger$ & 0.402\\
     & \textsc{Prompting}-Filter-\texttt{EIB} & 2.94 & ~~\cc\textbf{2.71}$\ddagger$ & \cc2.66 & ~~\cc1.97$\dagger$ & ~~\cc2.14$\ddagger$ & 0.422 \\
    \bottomrule
\end{tabular}
}
\caption{Human evaluation of explanation quality on two out-domain tasks, along with Krippendorff's $\alpha$ reported. \textsc{Prompting}-\texttt{EIB} and \textsc{Prompting}-Filter-\texttt{EIB} use the initial explanation candidates produced by \textsc{Prompting} and \textsc{Prompting}-Filter, respectively, as model inputs. \colorbox[RGB]{230, 240, 255}{Bluegrey chunk} denotes the observed improvements of *-\texttt{EIB} compared with from large-scale pretrained language model *. 
$\dagger$/$\ddagger$ results significantly outperform the results of corresponding pretrained language models * (sign test with $p\text{-value}< 0.05/0.01$).} 
%
\label{tab:overall_human_eval}
\end{table*}

\paragraph{Datasets}
We consider evaluating \texttt{EIB} on a universal setting and use two NLP tasks excluded from the training corpus \textsc{MixExpl} (\S\ref{sec:data_construct}) to analyze the explanation generalization abilities of \texttt{EIB}.
(i) ECQA~\citep{AggarwalMAKSG20} for commonsense question answering.
We formulate QA pairs into prompts to steer a large PLM, i.e., OPT-13B~\citep{opt}, and generate initial explanation candidates as input to \texttt{EIB}.
(ii) e-SNLI ~\citep{CamburuRLB18} for natural language inference where the premise, hypothesis, and inference label are packed into prompt input. 
Details of the dataset statistics are shown in Table~\ref{tab:stat_dataset}.

\paragraph{Baselines}
We compare \texttt{EIB} with the following baselines:
(i) \textsc{Supervised}. A supervised GPT-2 Small fine-tuned on target domain (i.e., ECQA and e-SNLI). 
(ii) \textsc{Prompting}. The prompt-based zero-shot learning framework with a PLM (OPT-13B).
(iii) \textsc{Prompting}-Filter. A trained acceptability filter on human binary judgments determines which of eight explanation candidates from PLM is plausible~\citep{WiegreffeHSRC2022}.
(iv) \textsc{BottleSum}. A reference-free summarization method~\citep{WestHBC19} using information bottleneck to extract highlight spans from a given paragraph (initial explanation candidates generated by PLM in this paper).

\paragraph{Training Details}
The backbone language models used in \texttt{EIB} are initialized from GPT-2 Small~\citep{radford2019language} with default parameters. 
During training, we use Adam optimizer~\citep{KingmaB14} with a learning rate of 5e-5. We train for 20 epochs with early stopping with mini-batches of size 32. 
For each explanation candidate, we average over 5 i.i.d. samples of compression distribution $\rv{t}$ to reduce the variance of the stochastic gradient where the compression weight $\beta$ is set to 1e-4 (Equation~\ref{eq:IB}). 
The dimension of each bottleneck vector $\mathbf{t}_i$ is 768 with a fixed length of 12. Explanations are generated by greedy decoding under the HuggingFace library~\citep{huggingface}

\subsection{\texttt{EIB} vs. {Baselines}}
\label{sec:main_human}

\paragraph{Overall Results}
Table~\ref{tab:overall_human_eval} shows the results.
We observe that \texttt{EIB} significantly outperforms \textsc{Prompting} and \textsc{Prompting}-Filter on the two testing tasks, and this superiority is consistent across different explanation attributes, especially for metrics factuality, sufficiency, and conciseness ($p < 0.05$, sign test).

Explanations polished by \texttt{EIB} are more concise and sufficient while maintaining good information coverage and quality, achieving over 44\% improvement on explanation refinement on the ECQA dataset, with a similar gain in the e-SNLI setting. 
The disparity in Grammar between the \textsc{Prompting}/\textsc{Prompting}-Filter methods and \texttt{EIB} is negligible. Slight deviations observed may be attributed to the comparatively concise predictions generated by \texttt{EIB}, resulting in a reduced number of errors.
\texttt{EIB} also substantially improves explanation quality over the edit-based method \textsc{BottleSum} for both tasks, while being more fluent, grammatical, and efficient where \textsc{EIB} (0.69 s/sample) infers much faster than \textsc{BottleSum} (55.01 s/sample). 

Notably, although \texttt{EIB} did not learn from any test domain datasets during training, it contains comparable performance with \textsc{Supervised} on explanation generation because of the knowledge retrieved from the gigantic PLM and the further refinement optimization towards sufficient and concise explanations.
We also evaluate the pair-wise comparisons between PLM and \texttt{EIB} on explanation generation and investigate the effectiveness of \texttt{EIB} on larger language models (i.e., GPT-3 175B). 
See Appendix~\ref{apx:hth_eval} and ~\ref{apx:gpt3} for more details.

Notably, the $\alpha$ values indicate that the level of agreement among annotators is not particularly high, a finding that is consistent with that of~\citet {WiegreffeHSRC2022},  likely due to the subjective nature of the task. 
Further information on evaluation quality control can be found in Appendix~\ref{apx:annotation}.

\begin{figure*}[!t]
\minipage{0.31\textwidth}
\includegraphics[width=\linewidth]{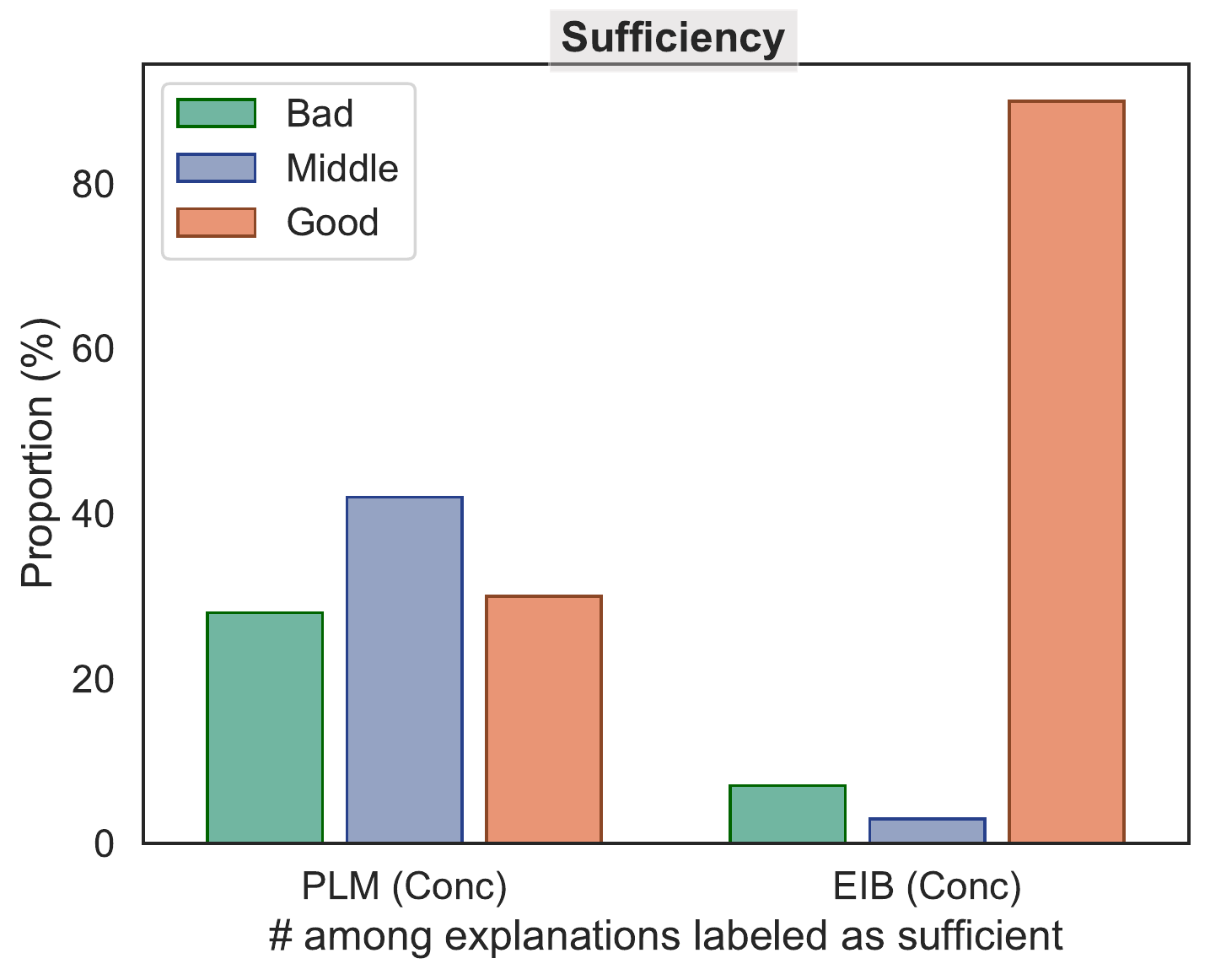}
  \label{fig:sufficiency}
\endminipage\hfill
\minipage{0.31\textwidth}

\includegraphics[width=\linewidth]{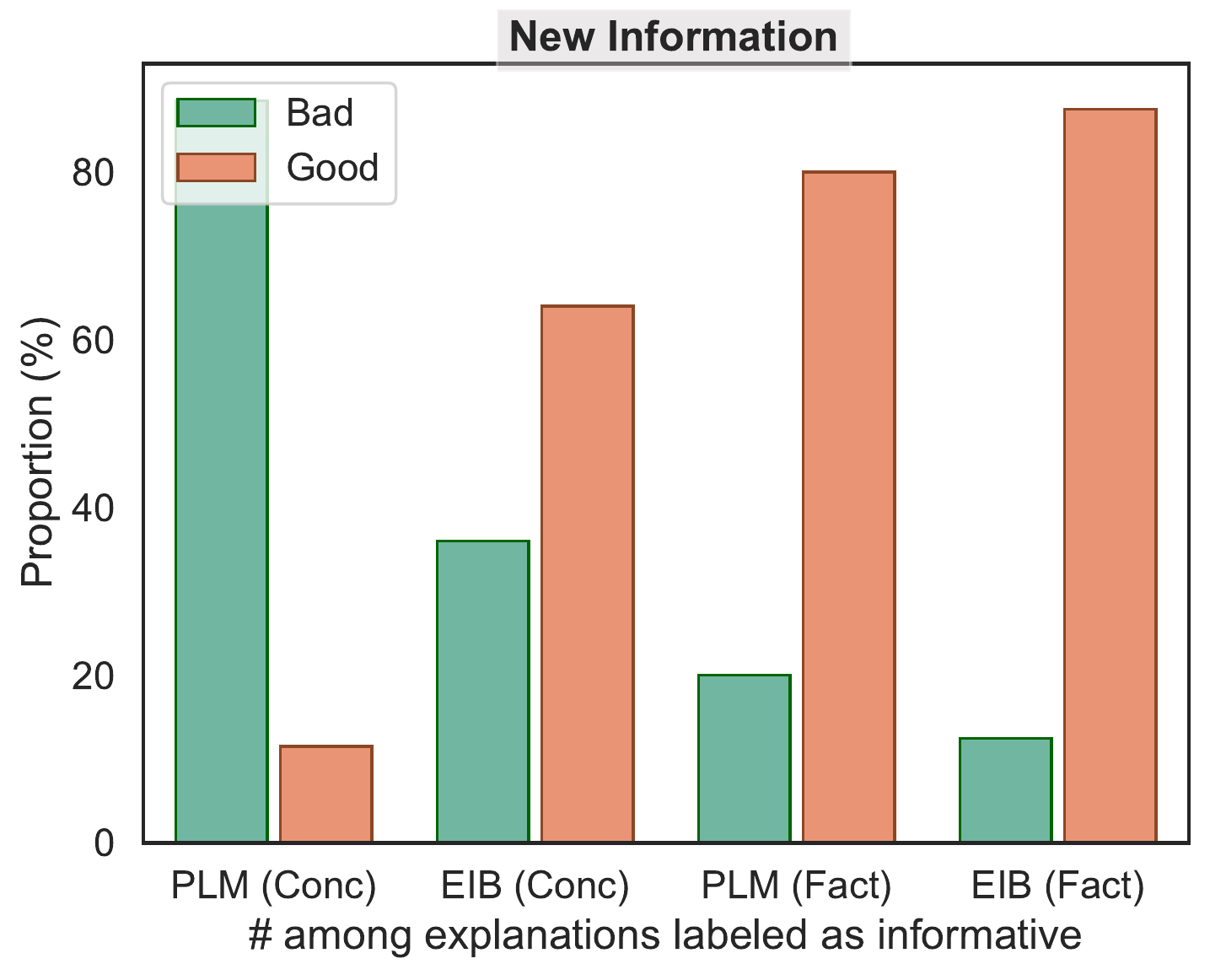}
  \label{fig:newinfo}
\endminipage\hfill
\minipage{0.31\textwidth}

\includegraphics[width=\linewidth]{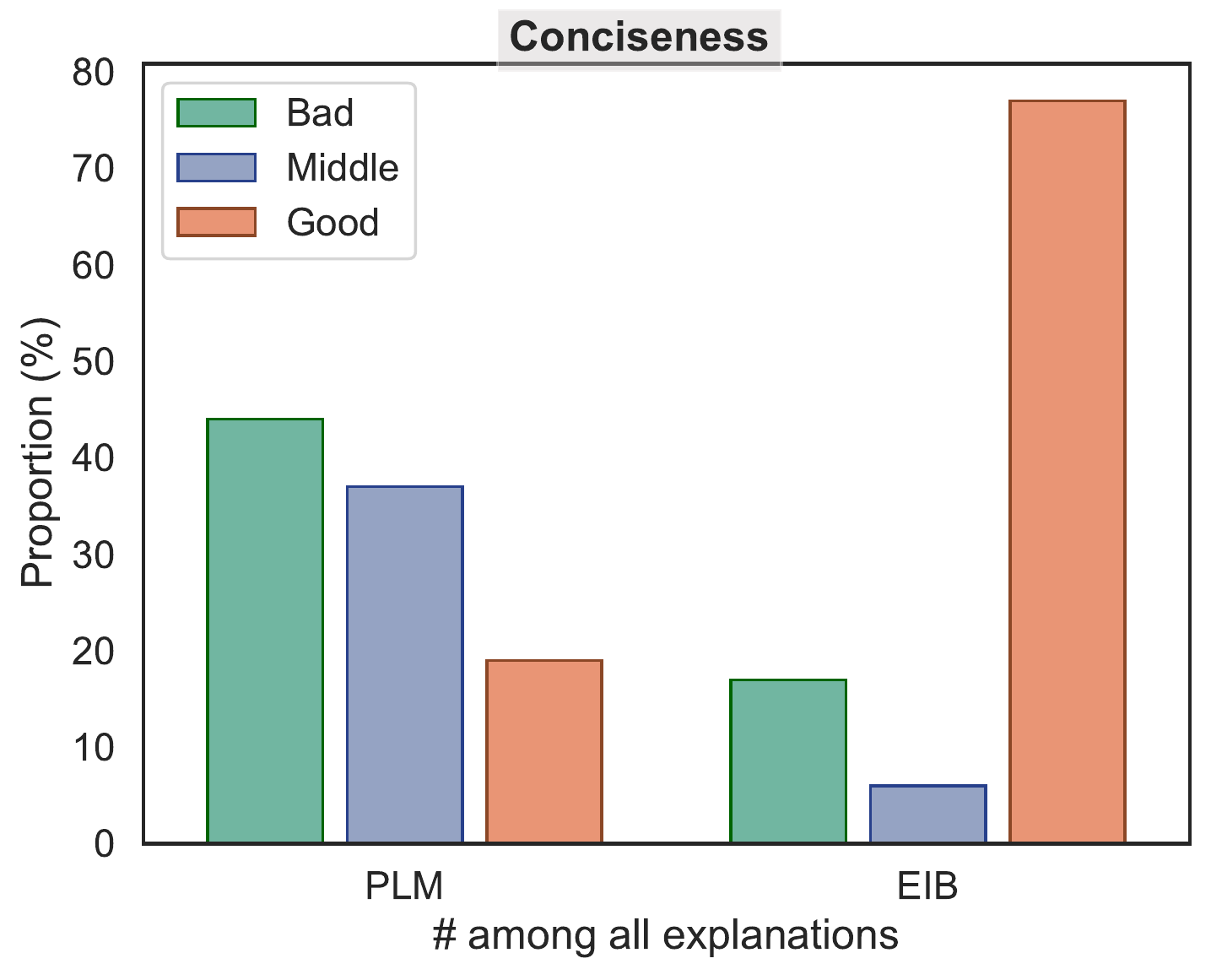}
  \label{fig:concise}
\endminipage 
\caption{
Comparison between \textsc{Prompting} and \texttt{EIB} under different explanation-level criteria. \texttt{EIB} outperforms the single-pass prompting method significantly with meaningful explanations while keeping reliable and concise.}
\label{fig:fine-grained-metrics}
\end{figure*}

\begin{table*}[!t]
\centering
\resizebox{1.98\columnwidth}{!}{
\begin{tabular}{llcccccccccc}
    \toprule
    \textbf{Dataset} & \textbf{Methods} 
    &\textbf{BERTScore}  &  \textbf{CIDEr} & \multicolumn{3}{c}{\textbf{BLEU}} &\multicolumn{2}{c}{\textbf{Distinct}}  &\multicolumn{2}{c}{\textbf{Novelty}} & \textbf{AVGLEN}\\ 
    \cmidrule(lr){5-7}\cmidrule(lr){8-9}\cmidrule(lr){10-11}
    & &   &  & \textbf{1} & \textbf{2} & \textbf{4} & \textbf{1} & \textbf{2} & \textbf{1} & \textbf{2} & \\
    \midrule
    ECQA &  \textsc{Supervised}  & 87.67 & 78.25 & 27.79 & 19.22 & 11.22 & 22.20 & 58.10 & 51.09 & 51.68  & 16.79 \\ 
    \midrule
    & \textsc{BottleSum} & 84.75 & 16.82 & 14.47 & ~~8.07 & \textbf{~~3.78} & 16.36 & 44.96 & 49.70 & 54.27 & 16.28 \\
    & \textsc{Prompting} &  84.38  & 14.48 & 14.31 & ~~7.57 & ~~3.15 & 11.45 & 34.37 & 46.87 & 54.72 & 27.47\\
    & \textsc{Prompting}-Filter & 85.35 & 17.10 & \textbf{15.52} & \textbf{~~8.10} & ~~3.39 & 13.14 & 47.49 & 54.35 & 61.44 & 27.22 \\
    & \textsc{Prompting}-\texttt{EIB} & ~~\cc85.02$\ddagger$ & ~~\cc16.76$\ddagger$ & 13.12 & ~~6.79 & ~~2.78 & ~~\cc14.12$\ddagger$ & ~~\cc37.71$\dagger$ & ~~\cc49.46$\dagger$ & ~~\cc56.95$\dagger$ & 15.46 \\
    & \textsc{Prompting}-Filter-\texttt{EIB}  & ~~\cc{\textbf{85.86}}$\ddagger$ & ~~\cc{\textbf{20.51}}$\ddagger$ & 15.25 & ~~7.92  & ~~3.19 & ~~\cc{\textbf{16.54}}$\ddagger$ & ~~\cc\textbf{48.44}$\ddagger$ & ~~\cc\textbf{55.10}$\ddagger$ & ~~\cc\textbf{61.60}$\dagger$ & 16.59 \\
    \midrule
    eSNLI & \textsc{Supervised} & 88.84 & 88.23 & 30.22 & 10.31 & 20.31 & ~~5.42 & 22.74 & 29.47 & 35.42 & 12.23 \\
    \midrule
    & \textsc{BottleSum}  & 85.95 & 38.02 & 20.97 & 13.17 & \textbf{~~6.01} &  ~~5.45 & \textbf{23.96} & 25.34 & 32.35 & 18.75 \\
    & \textsc{Prompting} & 85.83 & 17.23 & 16.99 & 10.32 & ~~4.49 & ~~3.60 & 15.61 & 27.09 & 36.24 & 27.65 \\
    & \textsc{Prompting}-Filter & 86.41 & 19.49 & 18.21 & 11.62 & ~~5.40 & ~~3.40 & 16.88 & 27.19 & 34.58 & 12.98\\
    & \textsc{Prompting}-\texttt{EIB} &  ~~\cc86.61$\ddagger$ & ~~\cc32.72$\ddagger$ &  ~~\cc20.96$\ddagger$ & ~~\cc11.77$\ddagger$ & ~~\cc~~4.83$\ddagger$ & ~~\cc~~5.52$\ddagger$ & ~~\cc20.30$\dagger$ & ~~\cc\textbf{32.03}$\dagger$ & ~~\cc\textbf{40.06}$\dagger$ & 13.78 \\
    & \textsc{Prompting}-Filter-\texttt{EIB}  &  ~~\cc\textbf{87.16}$\ddagger$ & \cc\textbf{42.88}$\ddagger$ &  \cc\textbf{22.30} & ~~\cc\textbf{13.52}$\ddagger$ & ~~\cc~~5.97$\ddagger$ &  ~~\cc\textbf{~~5.70}$\ddagger$ & ~~\cc22.65$\ddagger$ & ~~\cc30.85$\ddagger$ & ~~\cc37.01$\ddagger$ & 15.34 \\
    \bottomrule
\end{tabular}
}
\caption{Automatic evaluation of explanations generated by different models on the complete test splits of two datasets. Except for AVGLEN metric, other metric values are displayed in the percentage format.  Results that the \texttt{EIB} model outperforms its base PLM model are in \hl{greyblue}. 
$\dagger$, $\ddagger$ represent the significant improvement over the results of corresponding pretrained language models * with $p\text{-value}< 0.05/0.01$ respectively (sign test).
} 
\label{tab:overall_automatic_eval}
\end{table*}

\subsection{Fine-grained Explanation Quality}
\label{sec:fine-grained-quality}
We further analyze the \texttt{EIB}'s capacity to satisfy the semantic requirements of free-text explanations under three explanation-level evaluation features, new information, sufficiency, and conciseness.
Figure~\ref{fig:fine-grained-metrics} reports results on the ECQA dataset.

\paragraph{Sufficiency}
Among all sufficient explanations, \texttt{EIB} could achieve a better trade-off between sufficiency and conciseness, likely because of the optimization towards explanation refinement and polishing, pruning irrelevant information while attaining sample-relevance evidence.
For explanations labeled as ``introducing new information'' (middle figure), \texttt{EIB} significantly outperforms the prompting-based method with larger proportions of concise and factual explanations.
This indicates that \texttt{EIB} improves the quality of newly-introduced information in concise and convincing statements.

\paragraph{Conciseness}
We evaluate the main reasons causing explanations identified as ``redundant''. \textit{Bad} denotes copying the precedent context or repeating itself. \textit{Middle} represents containing off-topic content.
Compared to \textsc{Prompting}, the redundant issues could be largely alleviated by \texttt{EIB}, with a rising diversity proportion of abstract tokens that occurs in explanations, from 72.16\% to 85.24\%.

\subsection{Comparison on Automatic Metrics}
\label{sec:exp_ib}

\paragraph{Overall Results}
For comprehensive comparisons, we also investigate the performance of different methods on various automatic metrics. 
Results are shown in Table~\ref{tab:overall_automatic_eval}. 
The \textsc{Supervised} performs best among all methods. Our conjecture is that there are spurious correlations in test task datasets~\citep{KavumbaTO22}, e.g., for e-SNLI, golden explanations tend to use ``\textit{... a paraphrase of ...}'' to explain samples with ``\textit{entailment}'' labels. 
Among the unsupervised methods, we find that \texttt{EIB} improves generation qualities on most metrics over edit-based method (\textsc{BottleSum}) and prompting methods.
The improvement of \texttt{EIB} on vector-based metrics (BERTScore) and n-gram-based metrics (Distinct and Novelty) within a shorter length, leading to more sufficient and concise explanations.

\begin{table}[!t]
\centering
\resizebox{\columnwidth}{!}{
\begin{tabular}{llcccccccc}
    \toprule
    & \textbf{Methods} 
    &\textbf{BScore} & {\textbf{BLEU}} & {\textbf{Distinct}}  & {\textbf{Novelty}} & \textbf{AVGLEN}\\
    \midrule
    & \texttt{EIB} & 85.86  & ~~3.19 & 48.44 & 61.60 & 16.59\\
    \midrule
    &  w/o info preservation & 84.47  & ~~2.78 & 31.01 & 54.52 & 20.07\\
    &  w/o refinement & 84.44 & ~~1.88 & 19.47 & 50.76 & 23.17\\
    \bottomrule
\end{tabular}}
\caption{Ablation study on the effectiveness of information preservation objective and information bottleneck principle for ECQA dataset. We report on BERTScore, BLEU-4, Distinct-2, Novelty-2, and averaged length. 
} 
\label{tab:abla}
\end{table}

\begin{table}[!t]
\centering
\small
\begin{tabular}{p{0.93\linewidth}}
\toprule
\textbf{Premise:} The festivities of the latin celebration has brought many visitors and performers to the city. \\
\textbf{Hypothesis:} The city is completely devoid of people.  \\
\textbf{Label:} Contradiction \\
\midrule[0.03em]
\textbf{Human:} If the festivities brought many visitors and performers, it cannot be devoid of people. \\
\textbf{\textsc{Supervised}:} The Latin celebration is not entirely devoid of people. \\
\textbf{\textsc{BottleSum}:} People. The inference is that the city is full of people. The. \\ 
\midrule[0.03em]
\textbf{\textsc{Prompting}:} \textbf{\keepinfo{There are people}}.  The inference is \textbf{\keepinfo{that the city is full of people}}. \\
\textbf{\quad +\texttt{EIB}:} \textbf{\keepinfo{There are people.}} \textbf{\addinfo{The implication is}} \textbf{\keepinfo{that the city is full of people}}. \\
\midrule[0.03em]
\textbf{\textsc{Prompting}-Filter:} Because the city is completely devoid of people.  Now, let's look at the second example. \textbf{\keepinfo{Premise is the festivities of the latin celebration}}. \\
\textbf{\quad +\texttt{EIB}:} \textbf{\keepinfo{Premise is the}} \textbf{\addinfo{celebrations} \keepinfo{of the latin celebration}}. \textbf{\addinfo{People gather at the city's main square}}. \\
\bottomrule
\end{tabular}
\caption{Example from the e-SNLI dataset. Inherited information from the explanations of PLMs is colored in \textbf{\keepinfo{blue}}. Newly-added semantics are denoted in \textbf{\addinfo{orange}}. See Table~\ref{tab:more_case}, Appendix~\ref{apx:case} for additional examples.}  
\label{tab:case}
\end{table}

\paragraph{Effectiveness of Refinement}
The information bottleneck principle and information preservation objective (\S\ref{sec:t-info}) play key roles in refining imperfect explanation candidates into sufficient and concise ones, as shown in Table~\ref{tab:abla}.
The obvious decrease in reference-based metrics, such as BERTScore, demonstrates that the proposed information objective is beneficial for correct and concise explanations without losing on-topic information. 
To ablate the effect of the whole IB, we train a baseline on \textsc{MixExpl} without IB loss Equation~\ref{eq:IB} (w/o refinement), indicating that IB is very useful for generating sufficient and concise explanations.
A similar trend occurs in the e-SNLI dataset included in Appendix~\ref{apx:abla} Table~\ref{tab:apx_abla}.

\begin{figure}[!t]
     \centering
  \subfloat[BLEU]{\label{fig:bleuvshuman}\includegraphics[width=0.5\textwidth] {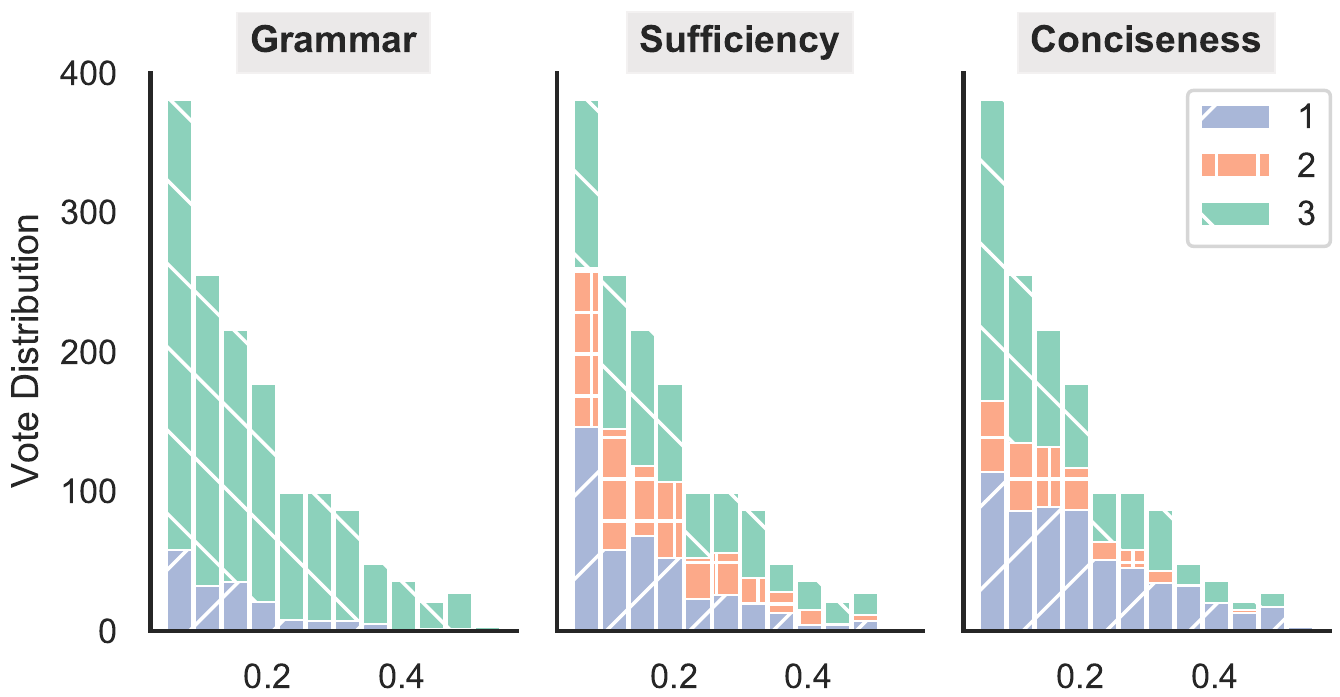}}\hspace{1em}
  \subfloat[BERTScore]{\label{fig:bertvshuman}\includegraphics[width=0.5\textwidth]{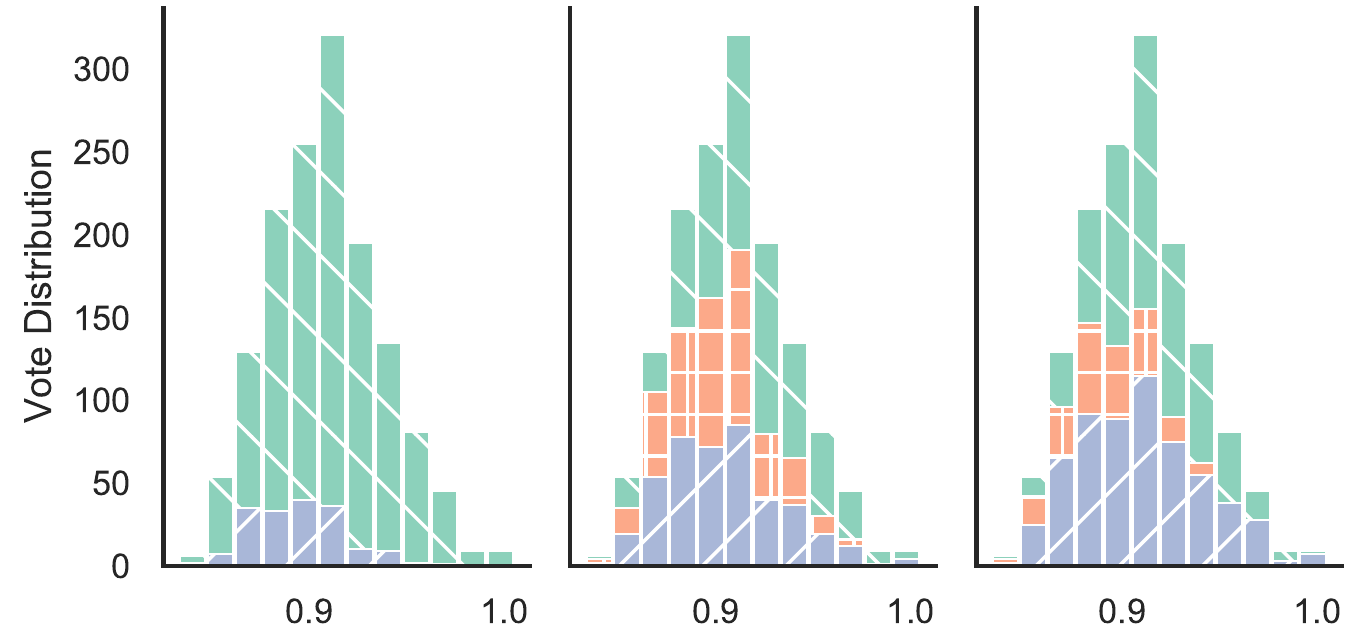}}
  \caption{The distribution of human evaluation scores across different ranges of automatic metrics, BLEU and BERTScore. Colour spans along the $y$-axis represent the human votes, ranging from 1 (worst) to 3 (best).} 
\label{fig:auto_vs_human}
\end{figure}

\subsection{Qualitative Analysis and Discussion}
\label{sec:qualitative_exp}

\paragraph{Cases}
Table~\ref{tab:case} displays an example of explanation generation for an NLI sample. The explanation generated by \texttt{EIB} is compelling enough as a more sufficient and concise version of the initial explanation candidates from prompting. Specifically, \texttt{EIB} corrects the explanation generated by \textsc{Prompting}-Filter, which initially contradicted the context, to be factual and sufficient.

\paragraph{Challenges}
The evaluation quality has a huge impact on designing explanation generation methods. 
We aim to answer ``\textit{are existing automatic metrics well-suited to evaluating zero-shot explanations?}''
Figure~\ref{fig:auto_vs_human} shows the agreement variation between the automatic and human metrics on the ECQA task.
On the language-level metric (grammar), both BLEU and BERTScore have strong consistency with human votes. 
However, for explanation-level metrics (sufficiency and conciseness), we can see an obvious disagreement between automatic and human metrics. The situation is worse for the simple $n$-gram matching BLEU. 
We see a noticeable percentage of explanations with low BLEU scores may acquire affirmation in human evaluation.
For BERTScore,
the issues have been alleviated, but they still exist.

Our finding is consistent with the recent works~\citep{Goyal22,holistic_eval22}.
Conventional evaluation difficulties in open-ended text generation also apply to explanation domains. 
Evaluating explanation generation, especially for unsupervised settings, will require a new framework distinct from conventional automatic metrics.

\section{Related Work}

Textual explanations in free-text forms are more expressive and generally more readable~\citep{RajaniMXS19}. 
Recent methods in free-text explanation generation could be divided into two types: supervised learning on labeled datasets~\citep{InoueTSBI21,ZhouJCLPR21,Scaffold22} and unsupervised learning with large-scale pre-trained language models (PLM)~\citep{Latcinnikb20,WiegreffeHSRC2022,Menicktmascgygim22,star22,palm22}. 
The success of zero-shot models~\citep{opt,gpt-3} drives research in a more reference-free way and saves annotation costs. 
A common strategy to encourage a PLM to produce explanations is to directly describe the input sample as context to the PLM, which has no guarantee for being supportive and organized explanations at one time~\citep{CamburuSMLB20,tan2021diversity,JungQABBBC2022,ye2022unreliability}.  
By contrast, \texttt{EIB} learns to distil task-relevance information from the initial explanations of PLM and regenerates sufficient and concise explanations with distant supervision from an automatically-constructed dataset.

Information bottleneck (IB) provides an information perspective to explain the performance of neural networks~\citep{Tishby2000}. IB measures the mutual information between random variables and is powerful, especially for unsupervised learning~\citep{OordliVinyals18}, which has been adapted in various NLP downstream applications~\citep{WestHBC19,paranjape-etal-2020-information,LiL20,ju-etal-2021-leveraging-information,referee22}, balancing a trade-off between task irrelevance and task objectives. 
We are interested in refining the unqualified explanation candidates into sufficient and concise ones with the guidance of the explained tasks by managing two IB objectives. To the best of our knowledge, we are the first to apply the information bottleneck principle to generate explanations that adhere to explanatory criteria.

\section{Conclusion}
Natural language explanations have attracted a lot of attention because free-text explanations are more expressive and generally more readable. However, the quality of machine-generated explanations still face challenges, e.g., inadequate evidences or redundancy expressions, even with large PLMs.  In this work, we propose to produce sufficient and concise explanations via the information bottleneck theory (IB), where explanations are regenerated by refining the single-pass outputs from PLM but keeping the information that supports the explained samples under a tradeoff between IB objectives. We automatically construct pseudo-parallel data for training \texttt{EIB} to autoregressively generate new explanations.
Experiments on two tasks show that \texttt{EIB} is effective for generating sufficient and concise explanations.  Besides, our extensive analysis shows that the current automatic evaluation for free-text explanation is extremely difficult, and persuasive evaluation frameworks are encouraged to compensate for conventional automatic metrics.

\section*{Limitations}
\paragraph{Extension to Varied Task Formats.}
In this work, we limit our experiments to generating free-text explanations given a complete task sample. In future work, we aim to extend our method over more diverse settings, e.g., controllable explanation generation or synergetic generation of both task prediction and explanation.
Besides, more work is needed to assess \texttt{EIB}’s robustness and generalization when applying it to diverse NLP domains. These domains may differ in sample type, topic, or even with different preferred explanation attributes. 

\paragraph{More lightweight Learning Paradigm.}
The performance of \texttt{EIB} is also tied to the quality of other systems or datasets, mainly the backbone language models and automatically constructed training corpus \textsc{MixExpl}.
The predictions of our method are also restricted by the capacity of the generator of \texttt{EIB}, where we use GPT2-small architecture as the decoding architecture.
This phenomenon may be remedied if we design specific interactions with larger PLM (e.g., in-context learning) and other sources for explanation-related knowledge distillation (e.g., logical composition). 
For example, designing more effective prompts to induce better explanation-related knowledge from PLM to relieve the training pressure.

\paragraph{Diverse Combination with PLMs.}
While our paper focuses on the issues of explanation generation given zero-shot prompting outputs, we think \texttt{EIB} is easy to extend to few-shot prompting baselines since single-pass generation without updating also belongs to the features of conventional few-shot settings. 
Currently \texttt{EIB} still needs parameter optimization. We think future work can explore more flexible plug-and-play methods to distill sufficient and concise explanations upon large PLM.

\paragraph{Evaluation Quality and Consistent.}
Quality estimation of the natural language explanation generation is largely dependent on human evaluation due to its open-ended characteristics. Current automatic evaluation metrics are not convincing and reliable when compared to human evaluation. However, reproducing the human evaluation results across different works may be difficult. This suggests that better automatic evaluation metrics are desperately needed for free-text explanation generation. We leave improving evaluation quality to future work.

\section*{Ethics Statement}
To comply with the ethics policy in ACL 2023, we analyze the potential ethical impact of our work, including transparency and privacy.

\paragraph{Transparency.} 
The motivation of our work is to generate free-text explanations that could sufficiently support the explained samples with concise expressions. We aim to provide faithful and trustworthy explanations in a human-readable way.

\paragraph{Privacy.} 
The language models and datasets we used are publicly available. Therefore, we do not harm the privacy of real users. 

Given the above demonstrations, we believe our research work will not violate ACL ethical code.




\normalem
\bibliography{custom}

\appendix

\section{Annotation Details}
\label{apx:annotation}

\begin{figure*}[!t]
    \centering
    \includegraphics[width=0.9\textwidth]{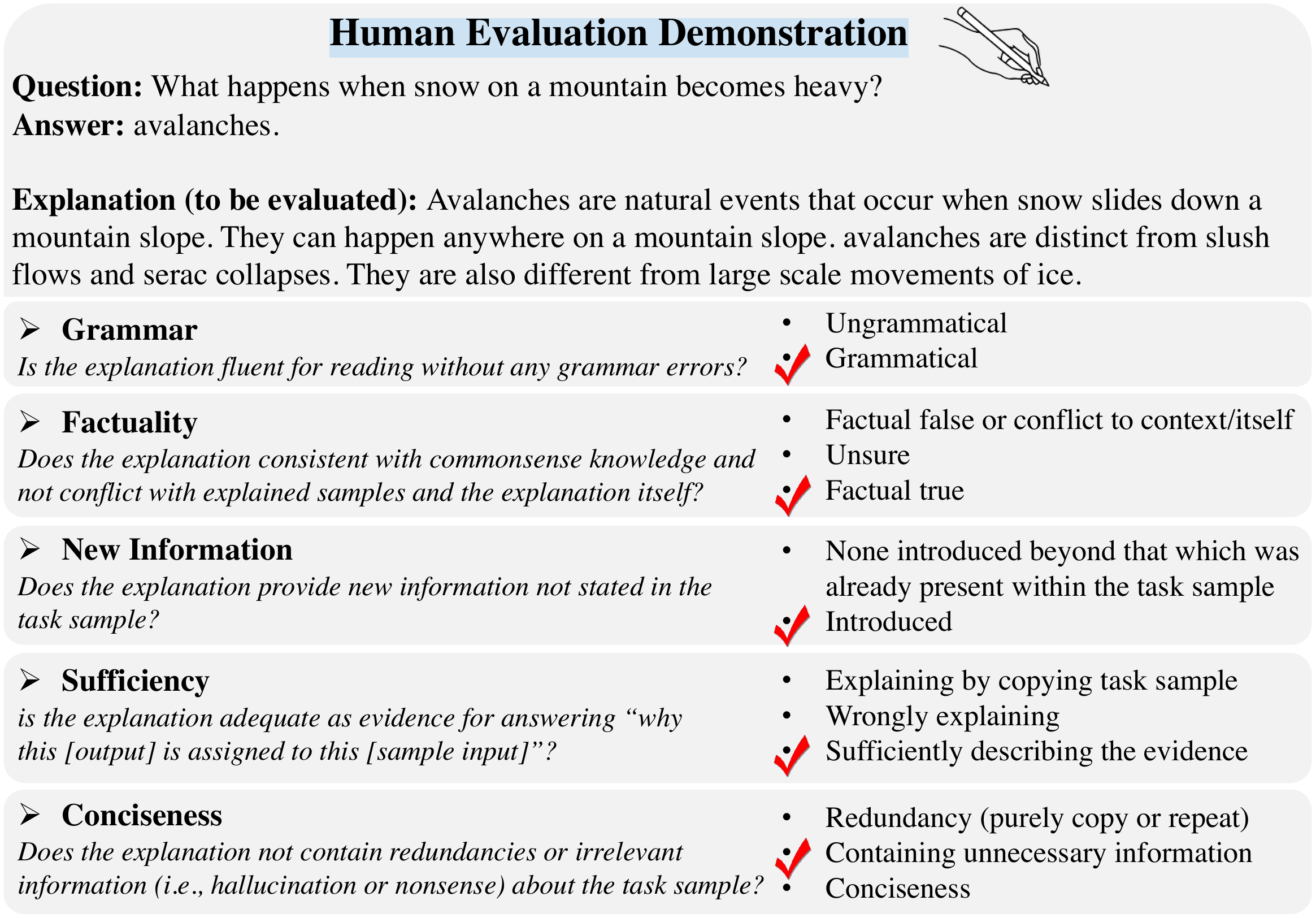}
    \caption{Demonstration of the head-by-head human evaluation pipeline. Given a task sample (e.g., QA) and an explanation candidate to be evaluated, annotators are required to evaluate the explanation candidate in 5 aspects. Two distinct options exist for Grammar and New Information metrics, while three-point scales are utilized for the evaluation of other metrics.}
    \label{fig:annotation_process}
\end{figure*}

\subsection{Human Evaluation Metrics}
Given a task sample and an explanation candidate to be evaluated, annotators are required to evaluate the explanation candidate in 5 axes: 
\begin{itemize}[wide=0\parindent,noitemsep,topsep=0em]
    \item \textbf{Grammar} (\textit{is the explanation fluent for reading without no grammar errors?} - yes or no). A natural-language explanation is at least fluent without grammatical mistakes. 
    \item \textbf{Factuality} (\textit{does the explanation consistent with commonsense knowledge and not conflict with explained samples and explanation itself?} -) Good explanations do not violate commonsense knowledge, not conflict with the established fact stated in the given sample or make self-contradiction.
    \item \textbf{New information} (\textit{does the explanation provide new information not stated in the task sample?} - ). During preliminary experiments, we found some explanations of PLMs tend to restate the given task sample declaratively. An explanation can be valid and factual (i.e., a restatement of the task sample), but not useful and vacuous~\citep{WiegreffeHSRC2022}.
    We expect a good explanation to be informative and meaningful, instead of a repeater.
    \item \textbf{Sufficiency} (\textit{is the explanation adequate as evidence for answering ``why this [output] is assigned to this [sample input]''?} -). Merely providing new information is not enough. If provided, the newly-introduced information should be compatible with the ``why question'' between the input and output of the task sample. Explanations are supposed to provide enough evidence to describe the relationship between sample input and output. 
    \item \textbf{Conciseness} (\textit{does the explanation not contain redundancies or irrelevant information?} - ) Explanations should be the selective and comprehensive reason over all possibilities, not to enumerate the complete set.
\end{itemize}

\subsection{Crowd-sourcing Instruction Details}

\paragraph{Head-by-head Evaluation of Table~\ref{tab:overall_human_eval}}
We show annotators the task sample (task sample input and output) and different explanations (six from models and one from human-written ground truth) and ask them to score each explanation along five evaluation attributes.
We instruct annotators to pretend the sample output is correct even if they disagree with it and judge the explanation based on the given output.
Specifically, for each choice of evaluated criteria, we detail the corresponding definitions to help explanation's error detection.
An illustration of the human annotation process is exemplified in Figure~\ref{fig:annotation_process}.
In practice, the annotation tasks were conducted online using shared Google files.

\paragraph{Head-to-head Evaluation of Table~\ref{tab:preference_eval}}

We present annotators with the task sample and instruct them to select which of two explanations best explains the task sample.
We ask them to ignore minor grammar and spelling mistakes such as improper upper casing.

\subsection{Quality Control}

We hire English native speakers as annotators from North America, to guarantee a high level of English proficiency among annotators. Annotators were pre-screened through a pilot qualification study. We showed them annotation requirements with three annotated examples by us (the authors) and require them to evaluate five representative samples.
On average, annotators took approximately five minutes to complete and perform a quick check for a single instance.
We pay them \$2 for every instance (6 explanations from models and 1 from human-written ground truth). 

We individually review submitted annotations of the qualification study and provide annotators with feedback to correct any misconceptions or confusion about the task. Annotators who performed well on the qualification study and demonstrated a comprehensive understanding of the task and annotation guidelines were permitted to participate in the main round of human evaluation. 
Finally, 3 annotators participated in the human evaluation.

Every few batches, we check to ensure the evaluation quality and time taken per annotator to avoid any annotator completing the tasks in an unreasonably quick time and containing inadvertent annotation errors. 
We maintained continuous communication with annotators throughout the human evaluation process to address queries and clarify intended behavior.
In order to track quality throughout evaluation, we compute inter-annotator agreement using Krippendorff’s $\alpha$ and hire new annotators to re-annotate if the disagreement is high among annotators ($\alpha < 0.3$).

Figures \ref{fig:annotation_guide1}-\ref{fig:annotation_guide3} show the annotation guidelines we provide for crowd annotators. We ask crowd annotators to read these guidelines before starting the qualification test. The annotators are required to contact us promptly if have any questions during the annotation.


\section{Additional Results}
\label{apx:more_analysis}

\subsection{Head-to-head Human Evaluations}
\label{apx:hth_eval}
We investigate whether the explanation regenerated by \texttt{EIB} better supports the explained task samples than the initial explanation candidates on the whole. 
We perform a head-to-head comparison of generations from prompting PLM (OPT-13B~\citep{opt}) vs. regenerations from \texttt{EIB}. 
We present three annotators with a task sample including input and output, and two explanations for the sample. We ask them to make a preferential selection by answering ```which explanation better explains the task sample?'''. Annotators are instructed to choose one option from a set of three alternatives: equivalence of the explanations, superiority of explanation 1, or superiority of explanation 2.

Results are shown in Table~\ref{tab:preference_eval}.
We find that, for both tasks, generations refined towards sufficiency and conciseness outperform the single-pass generations by prompting PLM. 
These results provide evidence that explanation refinement and regeneration are necessary for effectively explaining given samples because the special attributes of explanations are different from general language sentences.

\begin{table}[t!]
\centering
\resizebox{0.9\columnwidth}{!}{
\begin{tabular}{lccc}
    \toprule
    & \multicolumn{3}{c}{\textbf{Overall Explanation Preference (\%)}}  \\
    \cmidrule(lr){2-4}
    \textbf{Datasets} & \textbf{PLM} & \textbf{Tie} & \textbf{\texttt{EIB}}  \\
    \midrule
    ECQA  & 12.96 & 20.99 & 66.05\\
    e-SNLI  & ~~7.41 & 26.54 & 66.04 \\
    \bottomrule
\end{tabular}
}
\caption{A/B testing for explanations directly generated by the large-scale pre-trained language model (PLM) vs. additionally purified by EIB in two datasets, shown as \% preferences aggregated over 3 annotators.}
\label{tab:preference_eval}
\end{table}

\begin{table}[!t]
\centering
\resizebox{\columnwidth}{!}{
\begin{tabular}{llccccc}
    \toprule
    \textbf{Methods} & \textbf{New Info} & \textbf{Sufficiency} & \textbf{Conciseness} \\
    \midrule
    $\text{\textsc{Prompting}}_\text{\rm{13B}}$ & 2.69 & 2.02 & 1.73 \\
    \midrule[0.03em]
    $\text{\textsc{Prompting}}_\text{\rm{175B}}$ & 2.83 & 2.58 & 2.33 \\
    \texttt{EIB} & 2.83 & 2.70 & 2.79\\
    \bottomrule
\end{tabular}
}
\caption{Human evaluation of explanation quality for OPT (13B), GPT-3 (175B), and \texttt{EIB} on ECQA task.}
\label{tab:eib_vs_gpt3}
\end{table}

\begin{table}[!t]
\centering
\small
\begin{tabular}{p{0.93\linewidth}}
\toprule
\textbf{Question:} Playing baseball is a lot like any other sport, there is always a risk of what? \\
\textbf{Answer:} Injury. \\
\midrule[0.03em]
\textbf{Reference:} Sports is always a risk. \\
\midrule[0.03em]
\textbf{Few-shot Prompting:} Let's explain classification decisions.\textbackslash n\textbackslash nquestion: Where can someone view a county highway as a line?\textbackslash ncountry, from the sky, michigan, map, or cross counties by car? map\textbackslash nwhy? The only context in which an entire county road could be represented meaningfully as a single line would be on a map, e.g., a map of roads in the county, or a GPS application on a smartphone.\textbackslash n 
 \textit{\textcolor{textgrey}{... we omit the middle examples for simplicity}} \#\#\#\textbackslash nquestion: Joe bought a bunch of grapes because he likes to use them to top his what?\textbackslash ndeserts, make wine, bowl of fruit, shops, or sunshine? deserts\textbackslash nwhy? Grapes are sometimes a topping for desserts, e.g., fruit tarts.\textbackslash n\#\#\#\textbackslash nquestion: Playing baseball is a lot like any other sport, there is always a risk of what?\textbackslash ninjury, fun, sore muscles, happiness, or errors? injury\textbackslash nwhy?\\ 
\textbf{GPT-3:} \textbf{\keepinfo{Playing baseball is a lot like any other sport, there is always a risk}} of injury. Errors are not a risk in baseball. Happiness is not a risk in baseball. Sore muscles are not a risk in baseball. Fun is not a risk in baseball. \\  
\textbf{\quad +\texttt{EIB}:} \textbf{\keepinfo{Playing baseball is a lot like any other sport, there is always a risk}}.  \textbf{\addinfo{The risk of injury is a risk in baseball.  Sore muscles are a risk in baseball}}. \\
\bottomrule
\end{tabular}
\caption{Case study. GPT-3's prediction is provided by~\citet{WiegreffeHSRC2022}. Inherited information from the explanations of GPT-3 is colored in \textbf{\keepinfo{blue}}. Newly-added semantics are denoted in \textbf{\addinfo{orange}}. } 
\label{tab:gpt3_case}
\end{table}

\subsection{\texttt{EIB} vs. Few-shot GPT-3}
\label{apx:gpt3}
Furthermore, we want to investigate the effectiveness of \texttt{EIB} on larger sizes of PLM. We use the predicted explanations\footnote{\url{https://github.com/allenai/few_shot_explanations}} of GPT-3 Davinci with 175B reported by~\citet{WiegreffeHSRC2022}, where each prompt consists of 8-24 randomly selected human-written examples.
Annotators assess 100 samples of the ECQA dataset. The human evaluation results are shown in Table~\ref{tab:eib_vs_gpt3}. We can see that larger-scale GPT-3 (175B) performs much better than smaller OPT (13B) in producing meaningful and qualified explanations. \texttt{EIB} refines initial explanations generated by GPT-3 and could further improve the explanation quality.
\texttt{EIB} is much smaller than GPT-3. During inference \texttt{EIB} improves the explanation quality with a reduction of training FLOPs (46.420G) and model parameters (38.645M) by large orders of magnitude.

We also display an example in Table~\ref{tab:gpt3_case} for illustration. \texttt{EIB} keeps important contents of the initial explanation from GPT-3, abandons parallel sentences learned from the few-shot context, and further adds support to form a sufficient explanation.

\begin{table*}[!t]
\centering
\resizebox{1.95\columnwidth}{!}{
\begin{tabular}{llcccccccccc}
    \toprule
    \textbf{Datasets} & \textbf{Methods} 
    &\textbf{BERTScore}  &  \textbf{CIDEr} & \multicolumn{3}{c}{\textbf{BLEU}} &\multicolumn{2}{c}{\textbf{Distinct}}  &\multicolumn{2}{c}{\textbf{Novelty}} & \textbf{AVGLEN}\\ 
    \cmidrule(lr){5-7}\cmidrule(lr){8-9}\cmidrule(lr){10-11}
    & &   &  & \textbf{1} & \textbf{2} & \textbf{4} & \textbf{1} & \textbf{2} & \textbf{1} & \textbf{2} & \\
    \midrule
    ECQA 
    & \texttt{EIB} & 85.86 & 20.51 & 15.25 & ~~7.92  & ~~3.19 & 16.54 & 48.44 & 55.10 & 61.60 & 16.59\\
    \midrule
    &  w/o info preservation & 84.47 & 16.01 & 13.43 & ~~6.94 & ~~2.78 & 11.39 & 31.01 & 46.10 & 54.52 & 20.07\\
    &  w/o refinement & 84.44 & 12.76 & ~~9.70 & ~~4.95  & ~~1.88 & ~~7.14 & 19.47 & 40.69 & 50.76 & 23.17\\
    \midrule
    e-SNLI 
    & \texttt{EIB} & 87.16 & 42.88 &  22.30 & 13.52 & ~~5.97 &  ~~5.70 & 22.65 & 30.85 & 37.01 & 15.34 \\
    \midrule
    & w/o info preservation &  86.62 & 33.73 &  19.97 & 12.24 & ~~5.51 & ~~4.10 & 19.09 & 29.30 & 36.49 & 17.61 \\
    & w/o refinement & 86.46 & 33.79 &  19.53 & 11.89 & ~~5.31 &  ~~4.12 & 18.79 & 29.83 & 36.71  & 19.70 \\
    \bottomrule
\end{tabular}}
\caption{Ablation study for comparing the effectiveness of information preservation objective and information bottleneck principle on ECQA and e-SNLI dataset.} 
\label{tab:apx_abla}
\end{table*}

\begin{table*}[!t]
\centering
\resizebox{2\columnwidth}{!}{
\begin{tabular}{lccccccccccc}
    \toprule
    \textbf{\textsc{MixExpl}} & \textbf{BERTScore}  &  \textbf{CIDEr} & \multicolumn{3}{c}{\textbf{BLEU}} &\multicolumn{2}{c}{\textbf{Distinct}} &\multicolumn{2}{c}{\textbf{Novelty}} & \textbf{AVGLEN} \\
    & &  & \textbf{1} & \textbf{2} & \textbf{4}  & \textbf{1} & \textbf{2} & \textbf{1} & \textbf{2} &  \\
    \midrule
    Overall & 93.90 & 3.59 & 65.47 & 62.58 & 58.45 & 16.17 & 40.22 & 54.57 & 61.78 & 43.02 \\ 
    \midrule
    Science Exam QA~\citep{jansen2016s} & 92.99  & 2.81 & 50.76 & 48.25 & 44.55 & 10.28 & 22.08 & 43.81 & 56.38 & 63.76 \\
    Sen-Making~\citep{wang2019-make} & 94.39 & 4.43 & 45.49 & 42.86 & 37.36 & 28.84 & 51.77 & 62.13 & 70.81 & 13.84  \\
    LIAR-PLUS~\citep{alhindi2018} & 92.87 & 2.08 & 60.09 & 57.40 & 53.61 & 22.12 & 50.00 & 63.09 & 68.18 & 53.89 \\
    PubHealth~\citep{kotonya2020explainable} & 94.25 & 3.87 & 66.39 & 63.80 & 60.11 & 26.05 & 50.82 & 63.98 & 70.61 & 49.62  \\
    E-$\delta$-NLI~\citep{brahman2021learning} & 94.45 & 5.05 & 75.62 & 72.30 & 68.15 & 14.07 & 32.79 & 35.99 & 41.69 & 37.85  \\
    \bottomrule
\end{tabular}
}
\caption{The performance of \texttt{EIB} on the test set of \textsc{MixExpl}, as well as on the individual test sets of the five constituent tasks. Besides CIDEr and AVGLEN, other metrics are formatted into percentage values.
}
\label{tab:pretrain_task}
\end{table*}

\subsection{Ablation Study}
\label{apx:abla}
Results in Table~\ref{tab:apx_abla} show that the full model significantly improves the explanation quality across the different aspects, demonstrating the benefits of information bottleneck on explanation regeneration. Besides, our proposed information preservation loss ensures the usability of bottleneck representation with an obvious improvement on the reference-based metrics, e.g., for BERTScore, from 84.47 (w/o info preservation) to 85.86 (\texttt{EIB}).

\subsection{Performance on \textsc{MixExpl}}
\label{apx:mixexpl_test}
We also evaluate the performance of \texttt{EIB} on the test split of \textsc{MixExpl} and five trained tasks included in \textsc{MixExpl} to ensure the effectiveness of the training and generalization of the designed framework. Results are shown in Table~\ref{tab:pretrain_task}. The strong results on the test sets indicate the well-trained of \texttt{EIB} on the \textsc{MixExpl} corpus.

\section{Qualitative Examples}
\label{apx:examples}

\subsection{Prompting Format to PLM}
\label{apx:prompt_format}
When inference, the explanation candidates which are fed to \texttt{EIB} are prompted from large-scale pretrained language models (PLM). The prompting formats of test tasks (ECQA and e-SNLI) are illustrated in Table~\ref{tab:prompt_format}. 
We use OPT-13B as the PLM. The explanation candidates are generated by greedy decoding and top-$p$ sampling ($p$=0.9). For each example, we display one explanation candidate by greedy decoding and three candidates by top-$p$ sampling.

\subsection{Additional Cases} 
\label{apx:case}
More examples generated by PLM and \texttt{EIB} for ECQA and e-SNLI tasks are shown in Table~\ref{tab:more_case}.

\begin{table*}[!t]
\centering
\small
\begin{tabular}{p{0.9\linewidth}}
\toprule
\textbf{ECQA}~\citep{AggarwalMAKSG20} \\
\midrule[0.03em]
\textbf{Question:} The mother finished wrapping the very last gift, she then placed it under the what?  \\
\textbf{Answer:} Christmas tree.  \\
\textbf{Prompt \#1:} Let's explain question and answer. Question is the mother finished wrapping the very last gift, she then placed it under the what? Answer is Christmas tree \textcolor{newblue}{because} \\
\textbf{Prompt \#2:} LLet's explain question and answer. Question is the mother finished wrapping the very last gift, she then placed it under the what? Answer is Christmas tree. \textcolor{newblue}{Why?} \\
\textbf{Explanation Candidates:} \\
\quad (1) Greedy decoding: She was done. I'm not sure if you're being serious or not, but I'm going to assume you are. This is the only place where you can get a present the entire day. So the child could not see any presents the next day. So if you're still stuck, just choose another one of your friends! \\
\quad (2) Top-$p$ Sampling: This is the only place where you can get a present the entire day. So the child could not see any presents the next day. So if you're still stuck, just choose another one of your friends! \\
\quad (3) Top-$p$ Sampling: It would be the best place for a surprise. Congratulations to our winners! \\
\quad (4) Top-$p$ Sampling: Because she's a mother. She's got to do it. \\
\midrule
\textbf{e-SNLI}~\citep{CamburuRLB18} \\
\midrule[0.03em]
\textbf{Premise:} A statue at a museum that no one seems to be looking at.  \\
\textbf{Hypothesis:} Tons of people are gathered around the statue.  \\
\textbf{Label:} Contradiction \\
\textbf{Prompt \#1:} Let's explain a natural language inference. Premise is a statue at a museum that no one seems to be looking at. It is a contradiction to say tons of people are gathered around the statue \textcolor{newblue}{because} \\
\textbf{Prompt \#2:} Let's explain a natural language inference. Premise is a statue at a museum that no one seems to be looking at. It is a contradiction to say tons of people are gathered around the statue. \textcolor{newblue}{Why?} \\
\textbf{Explanation Candidates:} \\
\quad (1) Greedy decoding: It is not looking at anyone. The inference is that the statue is looking at someone. It is not a statue that everyone is looking at.  The same goes with the sentence ``The statue is not looking at everyone''. It is a contradiction to say tons of people are gathered around the statue. No one is looking at it. Inference is a process of finding out the truth of a statement. \\
\quad (2) Top-$p$ Sampling: It is not looking at anyone. The inference is that the statue is looking at someone. \\
\quad (3) Top-$p$ Sampling: No one is looking at it. Inference is a process of finding out the truth of a statement. \\
\quad (4) Top-$p$ Sampling: Because it is a contradiction. If tons of people are gathered around the statue then it would be obvious that they are looking at it. \\
\bottomrule
\end{tabular}
\caption{Prompting format for acquiring initial explanation candidates from large-scale pretrained language model using different prompting suffixes (because or Why?) and decoding strategies.} 
\label{tab:prompt_format}
\end{table*}

\begin{table*}[!t]
\centering
\small
\begin{tabular}{p{0.9\linewidth}}
\toprule
\textbf{ECQA}~\citep{AggarwalMAKSG20} \\
\midrule[0.03em]
\textbf{Question:} What is someone likely to call a bazaar in the us? \\
\textbf{Answer:} Festival.  \\
\textbf{PLM:} \textbf{\keepinfo{It's a celebration of the end of the harvest.  The bazaar is a place where people sell their goods.}} The word bazaar comes from the Arabic word bazara which means market. The word bazaar is used to describe a market place. \\
\textbf{\texttt{EIB}:} It's a celebration of the end of the harvest. The bazaar is a place where people sell their goods. \\
\midrule[0.03em]
\textbf{Question:} What type of feeling is performing for the first time likely to produce? \\
\textbf{Answer:} Anxiety.  \\
\textbf{PLM:} \textbf{\keepinfo{The performer is not sure how the audience will react.}}  Question: What is the difference between a performer who is nervous and a performer who is anxious?  Answer: \textbf{\keepinfo{A performer who is nervous is afraid of what the audience will think of him.}}  \\
\textbf{\texttt{EIB}:} The performer is not sure how the audience will react. A performer who is nervous is afraid of what the audience will think of him. \\
\midrule[0.03em]
\textbf{Question:} What type of feeling is performing for the first time likely to produce? \\
\textbf{Answer} Anxiety.  \\
\textbf{PLM:} They are \textbf{\keepinfo{competing in dangerous}} events. I think the question is more along the lines of ""why do people do this? \\
\textbf{\texttt{EIB}:} \textbf{\addinfo{People do this because they want to compete}}.  Competition is dangerous. \\
\midrule[0.03em]
\textbf{Question:} They were surprised that he enjoy rap, but he explained that every genre has what? \\
\textbf{Answer:} Good music.  \\
\textbf{PLM:} He thinks all music has good music i'd agree with that answer. You can make the argument that \textbf{\keepinfo{all music can be classified as good music or bad music by looking at the genre of the music}}. Rap music being so popular in today's. \\
\textbf{\texttt{EIB}:} \textbf{\addinfo{Rapper is a type of music}}.  All music can be classified as good music or bad music.\\
\midrule
\textbf{e-SNLI}~\citep{CamburuRLB18} \\
\midrule[0.03em]
\textbf{Premise:} A young boy is playing in the sandy water. \\
\textbf{Hypothesis:} A young boy is playing in the grass. \\
\textbf{Label:} Contradiction \\
\textbf{PLM:}  The boy is playing in the sand. The inference is that \textbf{\keepinfo{the boy is playing in the sand}}.  Premise is a young boy is playing in sand. \\
\textbf{\texttt{EIB}:}  The implication is that the boy is playing in the sand. \\
\midrule[0.03em]
\textbf{Premise:} A woman in a brown jacket and jeans is kneeling on a brick road while taking a picture of another woman posing. \\
\textbf{Hypothesis:} The woman is outside. \\
\textbf{Label:} Entailment \\
\textbf{PLM:} She is kneeling on the road. Premise is a woman in a brown jacket and jeans is kneeling on a brick road while taking a picture of another woman posing. \\
\textbf{\texttt{EIB}:}  \textbf{\addinfo{She is standing outside}}. \textbf{\addinfo{She is wearing a black jacket and black pants}}. \\
\midrule[0.03em]
\textbf{Premise:} A young couple riding on a bus with the boy's arm around the girl. \\
\textbf{Hypothesis:}  The couple are in their parent's car. \\
\textbf{Label:} Contradiction \\
\textbf{PLM:} \textbf{\keepinfo{Because the premise is that the couple are riding on a bus}} with the boy's arm around the girl. But the premise is not that the couple are riding in their parent's car. \\
\textbf{\texttt{EIB}:}  Because the premise is that the couple are riding in the bus \textbf{\addinfo{not their parent's car}}. \\
\bottomrule
\end{tabular}
\caption{Comparisons between generations from PLM and \texttt{EIB}. Inherited information of PLM is colored in \textbf{\keepinfo{blue}}. Newly-added contents are denoted in \textbf{\addinfo{orange}}.}
\label{tab:more_case}
\end{table*}

\begin{figure*}[!t]
    \centering
    \includegraphics[width=0.9\textwidth]{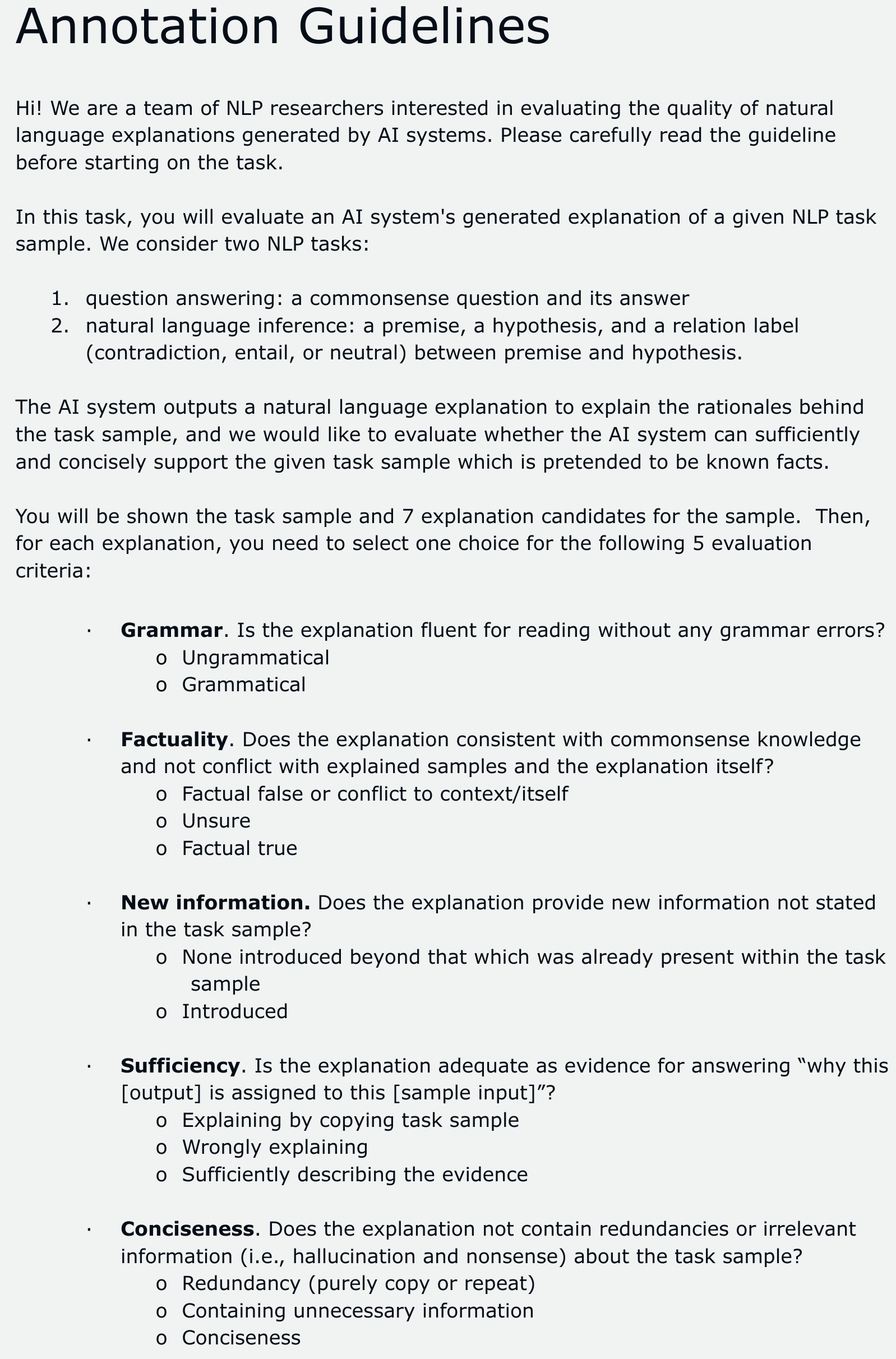}
    \caption{\qt{First page of the annotation guideline.}}
    \label{fig:annotation_guide1}
\end{figure*}

\begin{figure*}[!t]
    \centering
    \includegraphics[width=0.9\textwidth]{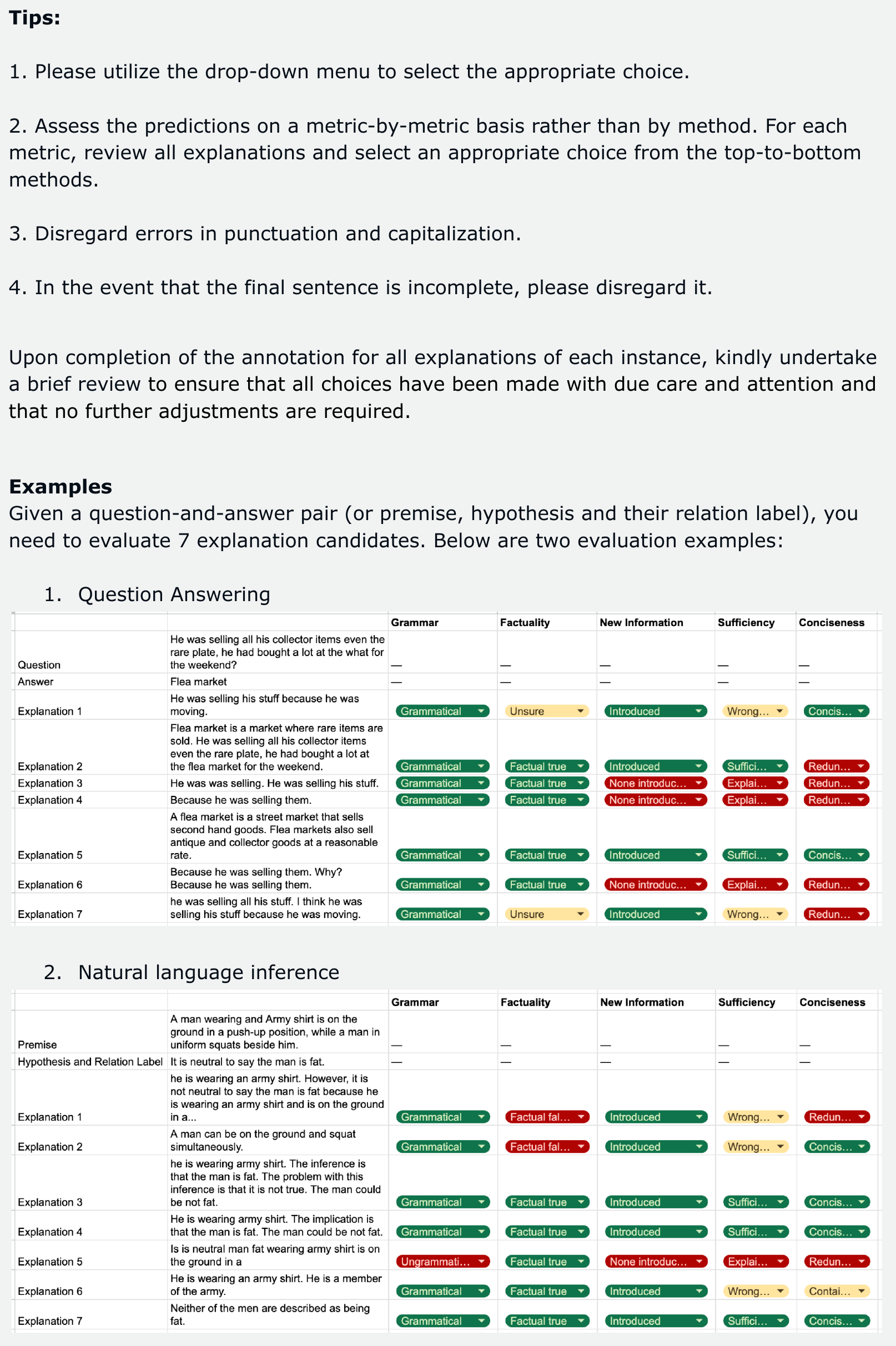}
    \caption{\qt{Second page of the annotation guideline.}}
    \label{fig:annotation_guide2}
\end{figure*}

\begin{figure*}[!t]
    \centering
    \includegraphics[width=0.85\textwidth]{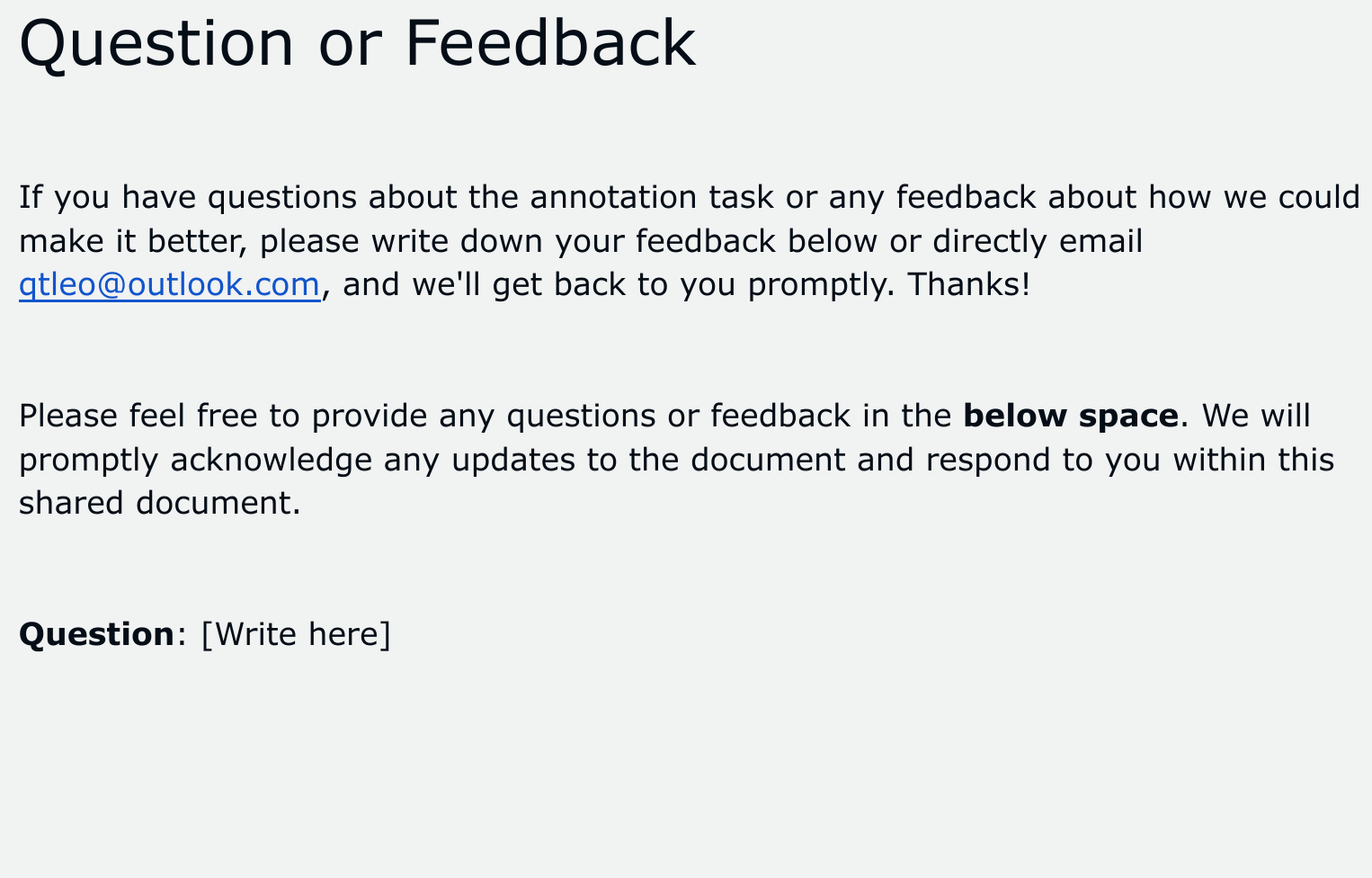}
    \caption{\qt{Third page of the annotation guideline.}}
    \label{fig:annotation_guide3}
\end{figure*}

\newpage
\end{document}